%% file: main.tex
\documentclass{article}
\usepackage[final]{colm2026_conference}

\usepackage[T1]{fontenc}
\usepackage[utf8]{inputenc}
\usepackage{inconsolata}
\usepackage{microtype}

\usepackage{amssymb}
\usepackage{latexsym}

\usepackage{array}
\usepackage{booktabs}
\usepackage{multirow}
\usepackage{multicol}
\usepackage{colortbl}
\usepackage{siunitx}

\usepackage{graphicx}
\usepackage{wrapfig}

\usepackage{listings}
\usepackage{adjustbox}
\usepackage[most]{tcolorbox}
\usepackage{fancybox}

\usepackage{soul}
\usepackage{enumitem}
\usepackage{url}

\usepackage{natbib}
\usepackage{hyperref}
\usepackage[capitalise,noabbrev]{cleveref}

\definecolor{darkblue}{rgb}{0, 0, 0.5}
\hypersetup{colorlinks=true, citecolor=darkblue, linkcolor=darkblue, urlcolor=darkblue}

\newcommand{\judgebenchmark}{\texttt{VERDICTS}}
\newcommand{\evalbenchmark}{\texttt{BFF-Bench}}
\newcommand{\cevalbenchmark}{\texttt{(C)MT-Bench}}
\newcommand{\symbolimg}[2][0.3cm]{%
  \ensuremath{\vcenter{\hbox{\includegraphics[height=#1]{#2}}}}%
}
\newcommand{\huggingface}{\symbolimg[0.3cm]{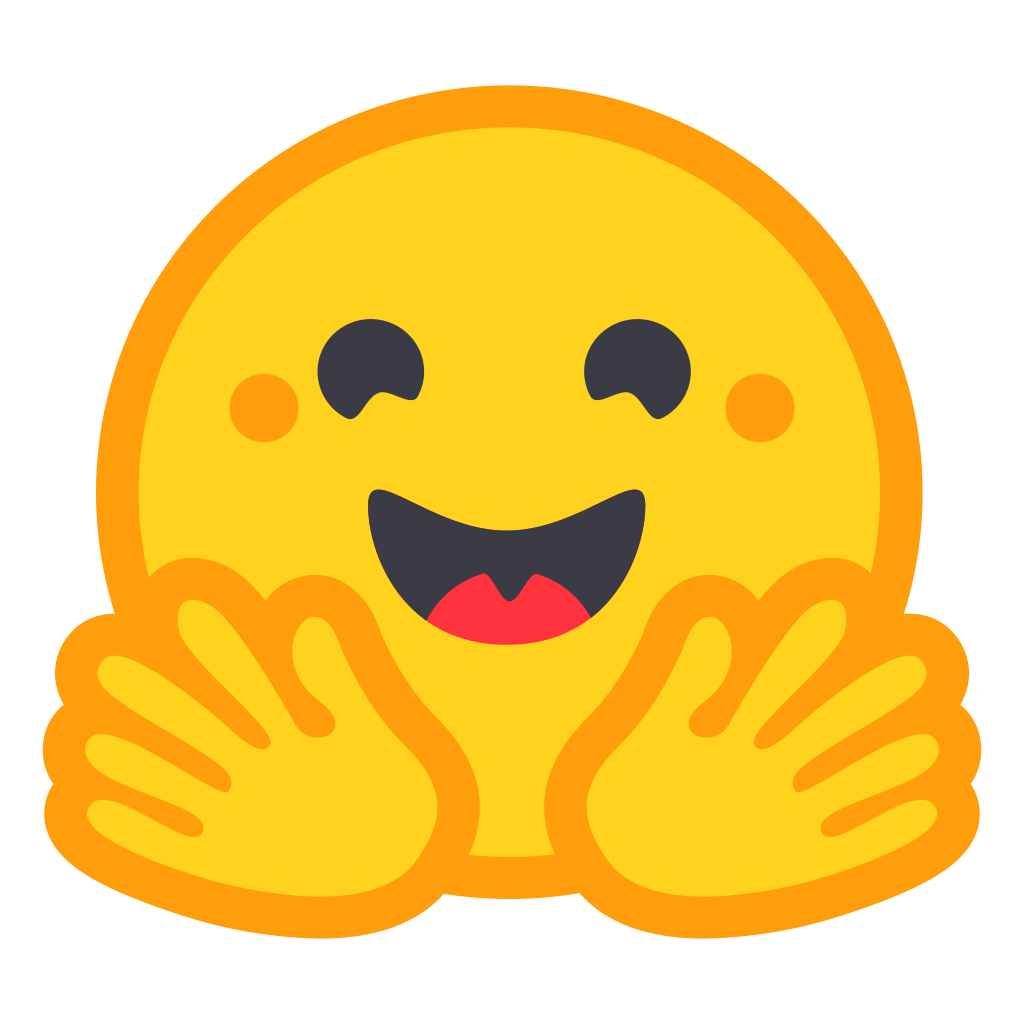}}
\newcommand{\R}{\cellcolor{red!30}}

\title{No Free Labels: Limitations of LLM-as-a-Judge Without \\Human Grounding}

\definecolor{providedbg}{RGB}{255, 235, 235}  %
\definecolor{providedborder}{RGB}{192, 57, 43}  %
\definecolor{oursbg}{RGB}{235, 255, 235}  %
\definecolor{oursborder}{RGB}{39, 174, 96}  %

\definecolor{k1}{HTML}{007694}
\definecolor{k2}{HTML}{B62699}
\definecolor{k3}{HTML}{00B9E8}
\definecolor{k4}{HTML}{0B008F}

\author{Michael Krumdick$^1$\;
Charles Lovering$^1$\;
Varshini Reddy$^1$\;
Seth Ebner$^1$\;
Chris Tanner$^{1,2}$
\\
$^1$Kensho Technologies \quad $^2$MIT \\
\texttt{\{michael.krumdick,chris.tanner\}@kensho.com} \\
}
  
\begin{document}

\maketitle

\begin{abstract}

Reliable evaluation of large language models (LLMs) is critical as their deployment rapidly expands, particularly in high-stakes domains. The LLM-as-a-Judge framework, which uses prompted LLMs to evaluate response quality, is appealing due to its scalability, low cost, and strong correlations with human stylistic preferences. However, it remains unclear how accurately these methods can assess response quality when correctness matters more than style. To address this gap, we introduce the Business and Finance Fundamentals Benchmark (\evalbenchmark), a dataset of 160 challenging questions and long-form responses authored by financial professionals. These experts subsequently evaluated the correctness of 1,200 responses generated by a diverse set of LLMs on both \evalbenchmark{} and a challenging subset of MT-Bench. With this expert-annotated dataset of judgments (\judgebenchmark{}), we analyze the agreement between a suite of automated grading methods and human experts. While we observe that LLM Judges are more reliable than other grading methods, our findings reveal a clear pattern in LLM Judge performance: when not provided with a correct reference, judges show high agreement with human experts \emph{only on questions the judges were able to correctly answer themselves}. We demonstrate that providing the judges with expert-written references largely mitigates this issue, highlighting the limits of using LLM-as-a-Judge without any form of human verification.

\end{abstract}
\begin{center}
\href{https://huggingface.co/collections/kensho/no-free-labels}{\huggingface~\texttt{kensho/no-free-labels}}
\end{center}

\section{Introduction}

For text generation tasks such as translation, summarization, and question answering, models provide free-form responses that are evaluated for some notion of overall quality \citep{Celikyilmaz2020EvaluationOT}. Because of the open-ended nature of the output space, automated evaluation methods often rely on lexical overlap heuristics that compare model outputs to trusted references (e.g., BLEU, \citealp{bleu}; ROUGE, \citealp{rouge}; and METEOR, \citealp{meteor}) or---more recently---on methods such as reward models \citep{ouyang2022training,kocmi-federmann-2023-large,fu-etal-2024-gptscore} and LLM-as-a-Judge \citep{mtbench,JudgeSurvey}. Using LLM-as-a-Judge for evaluation has become increasingly prominent due to its flexibility and high correlation with human preference. Benchmarks such as MT-Bench \citep{mtbench}, AlpacaEval2.0 \citep{alpacaeval2}, ArenaHardAuto \citep{arenahard}, and WildBench \citep{lin2024wildbench} are commonly used to evaluate the capabilities of frontier language models.

\input{figures/figure1}

The LLM-as-a-Judge framework is typically used in the following way: a model that we want to evaluate (the \textbf{candidate model}) generates responses to a set of benchmark questions. Then, to evaluate these responses, an LLM (the \textbf{judge model} or \textbf{LLM Judge}) is prompted to produce some kind of quality score from each response. The candidate model's response can be graded in isolation (\textbf{single grading}) or in comparison to an additional candidate model's response (\textbf{pairwise grading}). Each of the aforementioned benchmarks validates their use of LLM-as-a-Judge by measuring correlations with human preference annotations. Human preferences can naturally vary along many different axes, making it difficult to determine exactly why one response might be preferred to another \citep{DBLP:conf/icml/SorensenMFGMRYJ24,SOSBench,Szymanski2024ComparingCD}.

In some high-stakes domains such as business and finance, however, correctness outweighs other attributes \citep{bommasani2021opportunities,SOSBench,bizbench}. In these settings, accurately judging response quality hinges on identifying whether a response is correct, which itself can require complex reasoning and domain knowledge \citep{JudgeBench,mtbench}. In this work, we investigate the following question: \emph{Can LLMs make reliable judgments of correctness, and therefore quality, in these high-stakes domains?}

Our main contributions are as follows:
\begin{enumerate}
\item We create {\bf \evalbenchmark{}: a new, challenging conversational benchmark with a focus on correctness.} This business and finance themed benchmark contains 80 conversations spanning 160 questions with long-form reference answers created and verified by domain experts.
\item Using data from \evalbenchmark{} and MT-Bench, we create the \textbf{V}erified \textbf{E}xpert \textbf{R}esponse \textbf{D}ataset for \textbf{I}dentifying \textbf{C}orrectness over \textbf{T}ext \textbf{S}tyle, or {\bf \judgebenchmark{}: a dataset of expert judgments of correctness for 1,200 LLM responses} to evaluate the reliability of methods of pairwise and single grading.
\item We demonstrate that on \judgebenchmark{}, {\bf whether an LLM Judge agrees with human experts is strongly related to its ability to answer the underlying benchmark question and the correctness of the provided reference} (\cref{fig:one}).
\end{enumerate}

\input{sections/related_work}

\input{sample_bffbench}

\section{Datasets}
\label{sec:datasets}

To use as little synthetic data as possible in constructing our core benchmarks, we worked with human experts to create challenging questions with reference solutions, and annotate response correctness. A more detailed description of our annotators and annotation process can be found in \cref{sec:annotator_details}. We introduce a novel dataset (\evalbenchmark{}), a corrected subset of an existing dataset (\cevalbenchmark{}), and a set of human expert labels of model response correctness over these two datasets (\judgebenchmark{}).\footnote{All of these datasets are released with an Apache 2.0 license.}

\subsection{\evalbenchmark{}}

Our goal was to collect a benchmark that was conversational in nature but with unambiguous correctness. To have the same expertise available for both question creation and grading, we recruited a team of experts within a single high-stakes domain: business and finance. Although this limited the diversity of the underlying dataset, this tradeoff was necessary to scale both dataset creation and response grading.

We collected 80 two-turn conversational questions with gold reference answers, all written entirely from scratch by domain-experts. These questions are designed to be challenging and unambiguous, integrating both business and financial domain knowledge with quantitative reasoning. A sample question and reference answer are shown in \cref{fig:bffbench-sample}. (Untruncated examples are shown in \cref{fig:bffbench-sample-full,fig:bffbench-sample-full-2}.)

The question-response pairs were developed by a team of financial professionals. Each pair underwent an iterative quality control process and verification.
Questions were designed to have unambiguous answers to facilitate unambiguous correctness judgments.
Many require mathematical manipulation, from simple arithmetic to the application of financial formulas, whereas others test domain knowledge. Given their intended difficulty and technical scope, each question-response pair required approximately one hour to create, clean, and verify.

\subsection{\cevalbenchmark{}}

To increase the diversity of our judgment set, we collected data from the math and reasoning subsets (40 total questions) from MT-Bench. This subset, unlike MT-Bench's more open-ended categories,\footnote{For example, MT-Bench includes examples such as: ``Compose an engaging travel blog post about a recent trip to Hawaii, highlighting cultural experiences and must-see attractions.'' which admit multiple valid responses.} is evaluated with references by default in the MT-Bench pipeline and leads to responses that tend to be unambiguously correct or incorrect.

Inspired by \citet{inflectionai}, we reviewed the provided GPT-4-generated reference answers for these 40 questions and \emph{found that 15 (37.5\%) were either incorrect or inconsistent}.\footnote{By ``inconsistent'', we mean that the final answer is correct but the reasoning is invalid.} An example of one of these errors is provided in \cref{fig:mt-bench-wrong} in \cref{sec:cmt_examples}. We created correct references for each question by manually re-writing and verifying the existing references. These corrected references include the correct answer and a coherent reasoning chain, detailing all necessary sub-steps. We refer to this corrected subset as \cevalbenchmark{}. %

\subsection{\judgebenchmark{}}
We collected responses on \cevalbenchmark{} and \evalbenchmark{} from a diverse set of six models across various model families and sizes: \texttt{Gemma 2 2b} \citep{gemma2}, \texttt{Qwen 2.5 7b} \citep{qwen2.5}, \texttt{Phi-4} \citep{phi4}, \texttt{Yi 1.5 34B} \citep{yi15}, \texttt{Llama 3.3 70B} \citep{llama3}, and \texttt{GPT-4o} \citep{gpt4o}.\footnote{Our analysis entails domain-expert human annotations per response; extending to additional models requires new annotation rounds, so we prioritize diversity of model size/family over recency. We provide LLM-Judge-based results for additional models in \cref{sec:more-models}.} Each model was evaluated with a temperature of 0.7 using vLLM \citep{kwon2023efficient}, resulting in a total of 1,200 question-response pairs.\footnote{We used 8 80GB A100 GPUs for inference.}

For each question-response pair, we collected three annotations of correctness. For \evalbenchmark{}, the same financial professionals that created the questions served as annotators. Annotators received the human-written reference solution and could use any external resources except LLM-based chatbots. They were instructed to grade strictly: if any part was wrong, the entire response was marked wrong. Full guidelines are in \cref{sec:annotatorguidelines}. Dataset statistics are in \cref{tab:combined_dataset}.

\input{tables/datasets}

Annotators were instructed to select from ``Incorrect'', ``Correct'', and ``Not Sure.'' 54\% of the question-response pairs received unanimously consistent annotations, i.e., all three annotations were ``Incorrect'' or ``Correct''. 86\% of the questions had at most 1 annotation that differed from the other 2 annotations (e.g., 2 annotations were ``Incorrect'' and 1 was ``Not Sure''). 14\% of the questions had either no consensus or a consensus of ``Not Sure''. We deem these 14\% of examples to be ambiguous and exclude them from our experiments. %

\section{Experimental Setup}
\label{sec:experimental_setup}

We experiment on two judgment tasks derived from our \judgebenchmark{} dataset, measuring agreement with human annotators for both the standard LLM-as-a-Judge pipeline and a collection of baselines. The final dataset composition is in \cref{tab:combined_dataset}.

\subsection{Judgment Tasks}

For the {\bf single grading} task, the judge is given a question-response pair and must classify it as ``Correct'' or ``Incorrect''.  For the {\bf pairwise grading} task, the judge is given a question and two responses and must determine which is correct. We ensure that each pair has one response labeled ``Correct'' and one labeled ``Incorrect'' by the annotators. 

The single grading task matches how an LLM Judge would be used in practice for this type of judgment. 
In our analysis of the pairwise grading task, however, our setting represents a simplification of the real-world use-case by the removal of instances in which a model would have to distinguish correctness between two correct or two incorrect responses. If we were to consider pairs of responses that were both labeled correct or incorrect, the lack of an unambiguous preference label (is one correct response ``more correct'' than another?) would prevent straightforward analysis. Because of this simplified setting, the pairwise grading results should be interpreted only relatively (among judge models) rather than absolutely. Although measuring relative correctness can be a useful feature of pairwise judgment, this evaluation captures how well an LLM Judge can perform the most important pairwise distinction: between correct and incorrect responses.

We evaluate both tasks using Cohen's $\kappa$ \citep{CohenKappa}; an alternative version of the results using F1 are in \cref{tab:right-wrong-table-f1-v1}. More details on metrics and confidence intervals are in \cref{sec:confidenceintervals}.

\subsection{LLM-as-a-Judge}

Our five judges consist of the same set of models used as candidates. The only exception is \texttt{Gemma 2 2B}, which was used only as a candidate.\footnote{\texttt{Gemma 2 2B} was too inconsistent in its ability to produce the required judgment output format.} The judge model is allowed to use Chain-Of-Thought \citep{ChainOfThought} to first generate a set of reasoning steps before generating a final answer.  To reduce the overall variance in this analysis, we compute our judgments using self-consistency \citep{selfconsistency}, taking the majority vote over five separate inferences generated with a temperature of 0.7.\footnote{We selected 0.7 as the temperature since this was the value used in the original MT-Bench paper \citep{mtbench}.} We provide an ablation of the temperature and use of Chain-of-Thought in \cref{sec:judge-ablation}, but find neither to significantly impact performance on our tasks.

For the single grading case, the model is prompted to grade only the correctness of the model response. We include the full interaction history for the multi-turn questions within the judgment prompt. The model receives a similar prompt for the pairwise grading case, but with instructions to select the more correct response of the two provided. Following \citet{mtbench}, we run each pair twice, reversing the ordering of the references so as to account for any positional bias the judge model may have (e.g., always favoring the first model's response). If the judge model's selection differs between the two runs, we count it as a tie.

We use a version of the MT-Bench prompts modified to focus on correctness, which can be found in \cref{sec:judgeprompts}. To ensure our results are robust to the underlying prompt, we re-run our single grading experiments with a separate prompt that contains a rephrasing of the same instructions. These results are discussed in \cref{sec:prompt_sensitivity}, but we find no meaningful variation.

\subsection{Grading References}

Our main analysis concerns the impact of the provided reference on the judge's performance. Specifically, we look at three main types of provided references: (1) \textbf{\textit{None}}: no reference is provided within the LLM Judge's prompt, (2) \textbf{\textit{Self}}: the LLM Judge provides its own reference answer and (3) \textbf{\textit{Human}}: a human provides a manually created reference answer. In the \textit{None} case, the model is prompted to directly evaluate the provided response without a reference. For all other cases the prompt is identical except for the text of the reference.

In addition, we evaluate three types of baseline reference. \textit{Wrong} refers to human-edited versions of the \textit{Human} references that try to maintain the overall style but are incorrect. Each has some amount of inaccurate information or incorrect reasoning that changes the final answer to the question. Any correct response would then be incorrect with respect to these references. However, it should not impact the grading of responses that were already incorrect unless they happened to make the same error. \textit{Random} are randomly selected references from other questions. These references are non-informative without being actively misleading.

\subsection{Baselines}

We evaluate three automated grading baseline methods of response quality: lexical overlap metrics, embedding models, and reward models. Because these methods output quality scores, we focus our baseline evaluation on pairwise grading, labeling the response with the higher score as correct. The lexical overlap metrics capture how well-designed heuristics can perform on this task. We score each potential response with respect to the gold human-written reference. We evaluate BLEU \citep{bleu}, ROUGE \citep{rouge}, and METEOR \citep{meteor}. 

We convert an embedding model into a quality score by computing the cosine similarity between each response and the human-written reference. We select five high-performing, open-weight embedding models on the English split of MTEB \citep{mteb}.\footnote{MTEB does not require a public release of the model weights, so we were unable to access some of the top five models.} 

Reward models are LLM-based models explicitly trained to predict human preferences and are increasingly used within alignment pipelines. We compute the reward for each response directly, and select the response with the highest reward as being the ``Correct'' response. We select the five highest performing reward models on RewardBench \citep{RewardBench}. 

The full list of the baselines evaluated and their performance can be found in \cref{sec:baselines}.

\section{Results}

\input{figures/overall_performance}

\begin{table}[t]
\centering
\begin{minipage}{0.4\textwidth}
\centering
\begin{tabular}{lcc}
\toprule
Model & BFF & (C)MT \\
\midrule
GPT-4o & \textbf{68.1} & 75.0 \\
Llama 3.3 70b & 46.3 & \textbf{85.0} \\
Phi 4 (14b) & 51.9 & 77.5 \\
Qwen 2.5 7b & 46.3 & 67.5 \\
Yi 1.5 34b & 36.3 & 65.0 \\
Gemma 2 2b & 35.0 & 50.0 \\
\bottomrule
\end{tabular}
\caption{Correctness (\%) of candidate models on BFF-Bench and (C)MT-Bench as judged by human experts.}
\label{tab:human_rating}
\end{minipage}
\hfill
\begin{minipage}{0.56\textwidth}
\centering
\begin{tabular}{lc|cc}
\toprule
Reference & Overall & Correct & Incorrect \\
\midrule
\textit{Human} & $0.69_{\pm 0.06}$ & $0.74_{\pm 0.06}$ & $0.57_{\pm 0.08}$ \\
GPT-4o ($\checkmark$) & $0.61_{\pm 0.07}$ & $0.66_{\pm 0.07}$ & $0.52_{\pm 0.09}$ \\
\midrule
\textit{None} & $0.46_{\pm 0.08}$ & $0.56_{\pm 0.08}$ & $0.22_{\pm 0.09}$ \\
\textit{Random} & $0.50_{\pm 0.08}$ & $0.47_{\pm 0.08}$ & $0.09_{\pm 0.1}$ \\
\textit{Wrong} & $0.21_{\pm 0.06}$ & $0.22_{\pm 0.06}$ & $0.24_{\pm 0.09}$ \\
\bottomrule
\end{tabular}
\caption{Cohen's $\kappa$ with human annotators between different references, evaluated over the subset of benchmark questions that \texttt{GPT-4o} gets correct. \texttt{GPT-4o} ($\checkmark$) references are verified as correct.}
\label{tab:good-ref}
\end{minipage}
\end{table}

\cref{tab:human_rating} shows the human ratings of correctness for the six models as candidates on the two benchmarks. We find model performance to be generally correlated between the two benchmarks. Notably, BFF-Bench provided a greater challenge than (C)MT-Bench, with the best model achieving only 68.1\% correctness.

\cref{fig:overall-performance} shows the performance of LLM-as-a-Judge with different references compared to baseline methods on the single and pairwise judgment task. When provided with a human reference, even the worst LLM Judge outperforms the best baseline method. However, without the provided reference, some LLM Judges and reward models perform on par with heuristics.

The results between the single and pairwise grading do not directly reflect whether one setup is better than the other. The pairwise dataset was created to always have one correct and one incorrect response from two different candidate models. The large differences in response quality can make the judge model's grading relatively straightforward. As expected, we observed higher agreement with the human annotators in the pairwise grading setup than in the single grading setup.

\section{Analysis}

\paragraph{LLM-as-a-Judge with human-written references leads to the highest agreement with the human annotators.} Overall, providing a human-written reference improves judges' agreement with human annotators for all evaluated judges across both the single and pairwise cases. Notably, providing a human reference to a relatively small model (such as \texttt{Qwen 2.5 7B}) can yield better judgments than using a larger model without human references (such as \texttt{GPT-4o}) on the same set of responses.

\paragraph{Without a human-written reference, LLM Judges only achieve high agreement on questions they were able to answer as candidates.} For each question in the evaluation, we have a label indicating whether the judge model was able to answer that question correctly when used as a candidate model.\footnote{The label is the consensus judgment of the expert annotators.}  When we condition the judge's agreement on this label in \Cref{tab:right-wrong-table}, there are clear differences in judgment performance. For questions the judge model was unable to correctly answer, there are large differences in the agreement between using different reference types. This is especially the case for higher-performing judges. For example, when \texttt{GPT-4o} uses its own responses to questions as a reference (i.e., \textit{Self} response), the Cohen's $\kappa$ decreases from 0.86 to 0.16 in the pairwise case on questions \texttt{GPT-4o} could not answer. On questions the judge had answered correctly, providing human references still improves performance. However, while these differences are meaningful, they are much lower than when the model was incorrect. This is largely due to the higher judgment performance on correct instances. 

We include further analysis of this behavior in the appendix, demonstrating this effect is robust across both datasets used to create \judgebenchmark{} (\cref{sec:domain-generalization}), other models (\cref{sec:more-models}) and LLM Juries (\cref{sec:jury}).

\input{tables/correct_incorrect_table}

\paragraph{Different types of reference lead to different profiles of judge behavior.} One common judgment bias is self-preference \citep{panickssery2024llmevaluatorsrecognizefavor}, in which a model tends to overrate its own responses. To quantify this bias, we analyzed the different types of errors the LLM Judges were making. In this context, the false positive rate (FPR) represents the rate at which the model erroneously labels incorrect responses as being correct and the false negative rate (FNR) represents the rate at which the model erroneously labels correct responses as incorrect. A model with a strong self-preference bias would exhibit a higher FPR and a lower FNR when grading its own responses compared to when grading responses from other models.

\cref{fig:affinity_bias} displays the FPR and FNR aggregated over each judge with the \textit{Wrong}, \textit{Random}, \textit{None}, \textit{Self}, and \textit{Human} reference types on the single grading task.\footnote{Metrics per judge can be found in \cref{fig:affinity_per_judge_all}.} For every reference type, we see that on average models have a higher FPR when grading their own responses. The gap is particularly large when the model is provided with its own generated reference or no reference at all. Providing a human reference reduces the overall rate of both error types and the relative difference between a model's judgment of its own responses and those of others.

When the model is given a \textit{Wrong} reference, the FPR is low and the FNR is high, which suggests that the incorrect information contained in the reference leads judges to erroneously classify correct responses as incorrect.\footnote{Here, we mean erroneously in terms of the actual correctness of the underlying response. When given a \textit{Wrong} reference, the correct model behavior would be to rule everything incorrect, unless it happened to make the same error contained in the \textit{Wrong} reference.} However, this effect is not observed with the \textit{Random} reference. This evidence suggests that judges are sensitive to the correctness of the reference, and providing a slightly incorrect reference can in some cases be worse than providing a completely unrelated reference or no reference at all. 
\input{figures/affinity_bias}

Even when provided with a \textit{Wrong} reference that should make nearly every other response incorrect, LLM Judges were still frequently judging responses to be correct. We analyzed these judgments to understand why this occurred. Although most were simply reasoning errors (e.g. claiming the reference and the response agreed when they did not), we observed cases of a judge model identifying the reference as being incorrect and ``overruling'' the reference to correctly grade the response. An example of this ``overruling'' behavior is provided in \cref{sec:judge-refusal}.

\paragraph{Reference correctness matters more than being human-written.}
To better understand what makes human-written references lead to higher agreement with our human annotators, we compared performance across human-written gold references, a set of \texttt{GPT-4o} references judged by humans to be correct, a set of \textit{Wrong} references, and a set of \textit{Random} references. \cref{tab:good-ref} presents results for these  reference types across all five judges. The disparity observed in the previous section between human-written and synthetic references diminishes. There is also a significant gap between the \textit{Human} references and the \textit{Wrong} references. This suggests that the most important part of human-written references is their correctness, rather than any other superficial or stylistic attributes.

\section{Conclusion}
\label{sec:limitations}

{\bf We release \evalbenchmark{} and \judgebenchmark{}, two new datasets aimed at evaluating how well LLM-as-a-Judge grades response correctness in high-stakes domains.} These datasets enable the study of model performance both as candidates and as judges on difficult questions that require deep business and finance domain knowledge.

{\bf Using LLM-as-a-Judge does not alleviate the need for human oversight.} Our results demonstrate that LLM-as-a-Judge can lead to high agreement when: the LLM Judge  (1) can already answer the underlying question or (2) is provided with a correct reference. Having confidence in a judge model requires meeting at least one of these two conditions.

{\bf LLM-as-a-Judge performance is sensitive to the provided reference.} We observe large variations in performance when changing only the provided reference in the judgment prompt. Providing accurate references increases agreement with human annotators, while providing slightly incorrect references can be worse than providing no reference at all. The disparity in agreement is most pronounced for benchmark questions the judge model answers incorrectly. Improving agreement on such questions is key to advancing the reliability of evaluations of frontier models when using the LLM-as-a-Judge framework.

Therefore, {\bf we strongly recommend practitioners (at minimum) verify their reference responses.} While the scalability of LLM-as-a-Judge is one of the key attributes driving its continued adoption, our results suggest it may come at the cost of reliability. An LLM Judge might be able to consistently evaluate the stylistic merits of a response at scale, but there are clear limitations in the ability of these models to judge correctness.

The results in \cref{tab:good-ref} indicate that full human annotation for each reference might not be required, enabling some amount of scalability. Verifying responses generated by a model rather than manually writing and checking answers via annotators is sufficient in this setting. Although this verification process still requires some amount of human effort, our results suggest that this is unavoidable for developing trustworthy LLM-as-a-Judge pipelines. It is important to note that using verified references may improve the overall agreement metrics, but may also come at the cost of increased bias. Specifically, using a model's own references exacerbates the self-preference bias compared to providing no reference at all. We leave to future work a more in-depth analysis of the potential pitfalls of using verified, synthetic references and how best to enable both reliable and scalable LLM Judges.

\bibliographystyle{colm2026_conference}
\bibliography{main}

\appendix

\input{sections/appendix}

\end{document}

%% file: figures/figure1.tex
\begin{figure}[t]
    \centering
    \includegraphics[width=\linewidth]{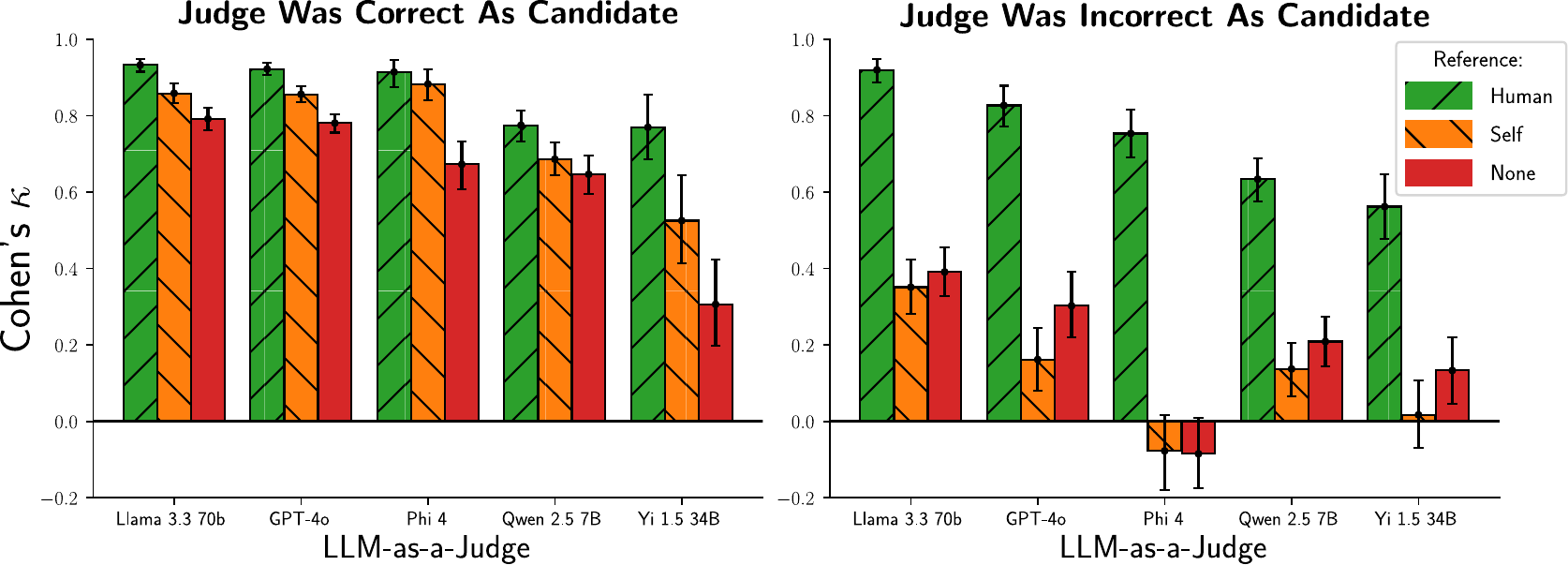}
    \caption{LLM Judge agreement with human annotators for the pairwise correctness judgment task. The left plot shows agreement when using a LLM-as-a-Judge on questions the LLM is able to answer correctly, whereas the right plot shows agreement on questions it answers incorrectly. \textit{None} means the LLM Judge is provided with no reference response, \textit{Self} means the LLM Judge is provided an unverified 
    reference generated by the judge model itself, and \textit{Human} means the LLM Judge is provided a human-written gold (correct) reference. When judge LLMs are not provided a correct reference, their agreement with human judgments decreases substantially for responses to questions they cannot correctly answer. However, providing a correct reference recovers much of this agreement.
    }
    \label{fig:one}
\end{figure}

%% file: sections/related_work.tex
\section{Related Work}

\textbf{Limitations of LLM-as-a-Judge.}
A growing body of work evaluates the shortcomings of LLM-as-a-Judge. Factors such as length, order, and tone of a response have all been shown to bias LLM Judges' evaluation \citep{mtbench,CognitiveBiases,SystematicEvaluationofLLMAsAJudge}. Benchmarks specifically designed to reveal such shortcomings have emphasized the challenges of developing reliable LLM Judges \citep{CognitiveBiases,JusticeOrPrejudice}. To mitigate such biases and enhance agreement with human annotators, various strategies have been proposed, including response-order randomization \citep{mtbench}, explicit debiasing against response length \citep{alpacaeval2}, ensemble-based “jury” models \citep{llmjuris}, rubric-driven evaluation frameworks \citep{hashemi-etal-2024-llm,rubrics}, prompt optimization techniques \citep{ZEPO}, pairwise preference optimization methods \citep{PAIRS}, and dynamically generated reference responses \citep{RevisEval}.

\textbf{LLM-as-a-Judge and Correctness.} Prior works have investigated the ability of LLMs to generally evaluate correctness or factuality \citep{min-etal-2023-factscore,song-etal-2024-veriscore,chen2023felmbenchmarkingfactualityevaluation}. Studies \citep{JudgingTheJudges,ReferenceGuidedVerdict,henkel2024largelanguagemodelsmake,badshah-etal-2025-clev} have demonstrated that LLMs can reach high agreement with human annotators when grading the correctness of responses to short-answer questions on existing datasets. \citet{SOSBench} specifically study how well standard LLM-as-a-Judge pipelines capture correctness and find that judge models often ignore response correctness in their quality assessments. \citet{JudgeBench} constructed a pairwise correctness judgment benchmark out of existing datasets, also finding poor performance.

\textbf{Expert-Informed LLM-as-a-Judge.} Many datasets incorporate human knowledge with verified reference responses for LLM evaluation \citep{bai-etal-2024-mt,malaviya-etal-2024-expertqa,chen-etal-2024-fintextqa,10.1609/aaai.v38i21.30362}.  \cite{recchia2025findtheflawsannotatederrorsdetecting} introduce a collection of five datasets with domain-specific questions, expert-verified long-form solutions, and annotated reasoning errors.  \citet{10.1145/3708359.3712091}  examines LLM-as-a-judge in the domains of dietetics and mental health by comparing GPT-4’s ratings to those of domain professionals, finding only moderate agreement on a dataset of 25 questions. In contrast, \judgebenchmark{} is the only dataset to contain questions, long-form answers, and judgments of correctness, all provided by the same domain experts.

%% file: sample_bffbench.tex
\begin{figure*}[t]
\centering
\footnotesize

\begin{tcolorbox}[
    boxsep=0.5pt,
    left=1pt,right=1pt,top=1pt,bottom=1pt,
    colback=blue!5!white,
    colframe=blue!40!black,
    before skip=1pt,
    after skip=1pt,
    sharp corners=downhill
]
\textbf{Question One:} What is the IFRS 16 treatment of short term or low value leased assets? What are the restrictions?
\end{tcolorbox}

\begin{tcolorbox}[
    boxsep=0.5pt,
    left=1pt,right=1pt,top=1pt,bottom=1pt,
    colback=green!2!white,
    colframe=green!25!black,
    before skip=1pt,
    after skip=1pt,
    sharp corners=downhill
]
\textbf{Gold Response:} IFRS 16 Leases permit a simplified treatment for assets with a lease period of 12 months or less, or of low value. Although the standard does not give a numerical definition of \dots
\end{tcolorbox}

\begin{tcolorbox}[
    boxsep=0.5pt,
    left=1pt,right=1pt,top=1pt,bottom=1pt,
    colback=blue!5!white,
    colframe=blue!40!black,
    before skip=1pt,
    after skip=1pt,
    sharp corners=downhill,
]
\textbf{Question Two:} On April 1st 2023, Abigail acquired telephones under a two-year lease agreement. The terms of the lease require an initial payment of \$3,000, followed by a payment of \$9,000 on March 31st 2024 and an additional \$9,000 on March 31st 2025. Show the impact of this lease arrangement on Abigail's financial statements for the year ended 31 December 2023 under IFRS 16?
\end{tcolorbox}

\begin{tcolorbox}[
    boxsep=0.5pt,
    left=1pt,right=1pt,top=1pt,bottom=1pt,
    colback=green!2!white,
    colframe=green!25!black,
    before skip=1pt,
    after skip=1pt,
    sharp corners=downhill,
]
\textbf{Gold Response:}
Annual lease rental expense $= (3000 + 9000 + 9000)/2 = 10500$ per annum.
Expense to 31 December 2023 $= 10500 \times \frac{9}{12} = 7875$.
Accrued expense $= 7875 - 3000 = 4875$.

The expense in this period of \$7,875 is not the same as the payment of \$3,000, so we need to accrue an additional expense of \$4,875.
\end{tcolorbox}

\caption{Truncated example of a question from BFF-Bench. For each question, we include a human-written gold answer that contains the final answer and a complete reasoning chain. The full version of this example is shown in \cref{fig:bffbench-sample-full} in \cref{sec:bff-bench-examples}.}
\label{fig:bffbench-sample}
\end{figure*}

%% file: tables/datasets.tex
\begin{table*}[t]
\centering

\begin{tabular}{l|cc|cc|cc}
\toprule
& \multicolumn{2}{c|}{\textbf{Benchmark}} & \multicolumn{2}{c|}{\textbf{\judgebenchmark{}}}  & \multicolumn{2}{c}{\textbf{Grading}}\\
\textbf{Dataset} & \emph{Convs.} & \emph{Turns} & \emph{Responses} & \emph{Annotations} & \emph{Single} & \emph{Pairwise} \\
\midrule
\evalbenchmark{} & 80 & 160 & 960 & 2880 & 808 & 442 \\
\cevalbenchmark{} & 20 & 40 & 240 & 720 & 228 & 162 \\
\midrule
\textbf{Total} & \textbf{100} & \textbf{200} & \textbf{1200} & \textbf{3600} & \textbf{1036} & \textbf{604} \\
\bottomrule
\end{tabular}

\caption{Dataset statistics for the benchmarks and the response grading dataset. The \emph{Single} and \emph{Pairwise} columns denote the number of single grading and pairwise grading examples after filtering for annotator consensus.}
\label{tab:combined_dataset}
\end{table*}

%% file: figures/overall_performance.tex
\begin{figure}[t]
    \centering
    \includegraphics[width=\textwidth]{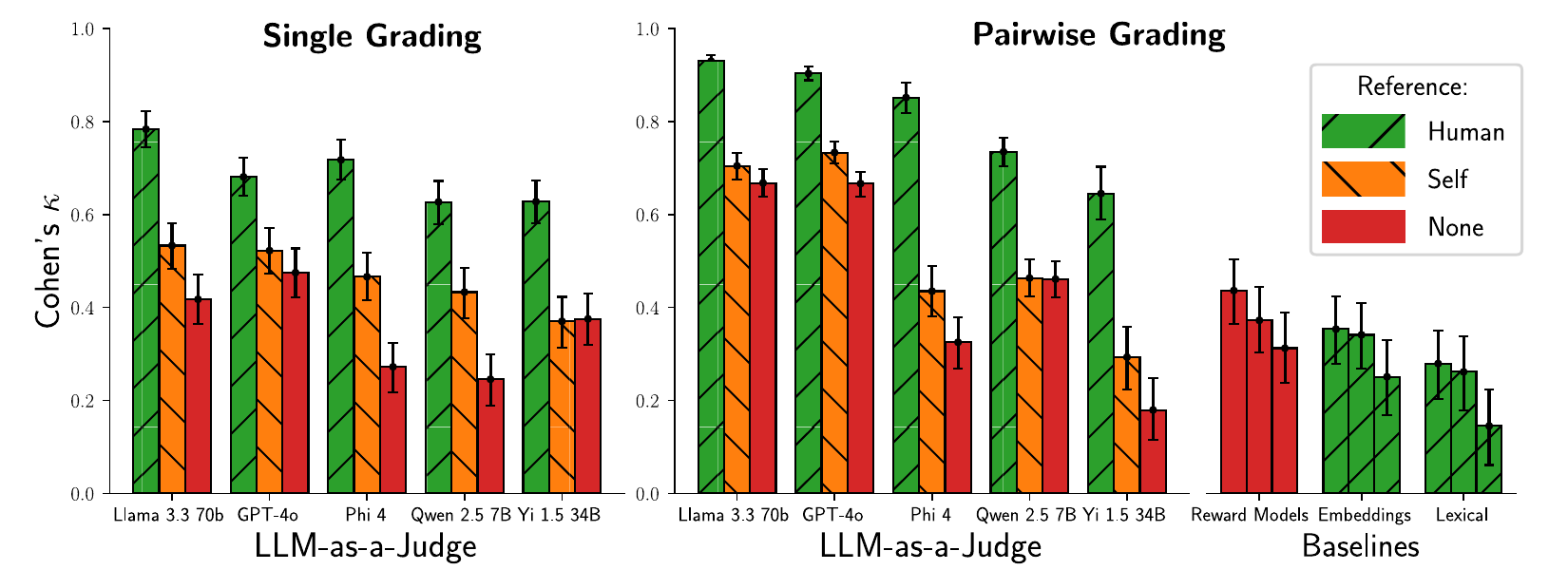}
    \caption{LLM Judge agreement (Cohen's $\kappa$) with human annotators across the pairwise and single grading cases for the three types of reference (human-written, judge-generated, or no reference). For every judge model in every setting, we observe an increase in agreement with the human annotators when using domain expert-written references.}
    \label{fig:overall-performance}
\end{figure}

%% file: tables/correct_incorrect_table.tex
\begin{table*}[t!]
    \centering
    \resizebox{\textwidth}{!}{
    
    \begin{tabular}{lc|ccc|ccc}
        \toprule
       & \textbf{As Candidate,} & \multicolumn{3}{c|}{\textbf{Pairwise Grading}} & \multicolumn{3}{c}{\textbf{Single Grading}} \\
         \textbf{Judge}  & \textbf{Judge was:} & \textit{None} & \textit{Self} & \textit{Human} & \textit{None} & \textit{Self} & \textit{Human} \\
        \midrule
\multirow{2}{*}{GPT-4o} & Correct & $0.78_{\pm 0.02 }$ & $0.86_{\pm 0.02 }$ & $\textbf{0.92}_{\pm 0.02 }$ & $0.46_{\pm 0.07 }$ & $0.52_{\pm 0.06 }$ & \R $0.59_{\pm 0.06 }$\\
& Incorrect & \R $0.30_{\pm 0.09 }$ & \R $0.16_{\pm 0.08 }$ & \R $\textbf{0.83}_{\pm 0.06 }$ & \R $0.16_{\pm 0.14 }$ & \R $0.13_{\pm 0.14 }$ & $\textbf{0.81}_{\pm 0.13 }$\\
\midrule
\multirow{2}{*}{Llama 3.3 70b} & Correct & $0.79_{\pm 0.03 }$ & $0.86_{\pm 0.03 }$ & $\textbf{0.93}_{\pm 0.02 }$ & $0.61_{\pm 0.08 }$ & $0.54_{\pm 0.08 }$ & $0.73_{\pm 0.07 }$\\
& Incorrect & \R $0.39_{\pm 0.06 }$ & \R $0.35_{\pm 0.07 }$ & $\textbf{0.92}_{\pm 0.03 }$ & \R $0.13_{\pm 0.06 }$ & \R $0.23_{\pm 0.10 }$ & $\textbf{0.69}_{\pm 0.09 }$\\
\midrule
\multirow{2}{*}{Phi 4} & Correct & $0.67_{\pm 0.07 }$ & $0.88_{\pm 0.04 }$ & $0.91_{\pm 0.04 }$ & $0.48_{\pm 0.09 }$ & $0.47_{\pm 0.07 }$ & $\textbf{0.67}_{\pm 0.07 }$\\
& Incorrect & \R $-0.08_{\pm 0.09 }$ & \R $-0.08_{\pm 0.10 }$ & \R $\textbf{0.75}_{\pm 0.06 }$ & \R $0.03_{\pm 0.05 }$ & \R $0.02_{\pm 0.12 }$ & $\textbf{0.56}_{\pm 0.12 }$\\
\midrule
\multirow{2}{*}{Qwen 2.5 7B} & Correct & $0.65_{\pm 0.05 }$ & $0.69_{\pm 0.04 }$ & $\textbf{0.78}_{\pm 0.04 }$ & $0.31_{\pm 0.11 }$ & $0.37_{\pm 0.09 }$ & $0.54_{\pm 0.09 }$\\
& Incorrect & \R $0.21_{\pm 0.07 }$ & \R $0.14_{\pm 0.07 }$ & \R $\textbf{0.63}_{\pm 0.06 }$ & \R $0.08_{\pm 0.07 }$ & $0.21_{\pm 0.10 }$ & $\textbf{0.56}_{\pm 0.08 }$\\
\midrule
\multirow{2}{*}{Yi 1.5 34B} & Correct & $0.31_{\pm 0.12 }$ & $0.53_{\pm 0.12 }$ & $\textbf{0.77}_{\pm 0.09 }$ & $0.23_{\pm 0.11 }$ & $0.29_{\pm 0.08 }$ & \R $0.43_{\pm 0.09 }$\\
& Incorrect & $0.13_{\pm 0.09 }$ & \R $0.02_{\pm 0.09 }$ & \R $\textbf{0.56}_{\pm 0.08 }$ & $0.18_{\pm 0.09 }$ & $0.14_{\pm 0.10 }$ & $\textbf{0.65}_{\pm 0.08 }$\\

        \bottomrule
    \end{tabular}
    }
    \vspace{0.5em}
        \caption{Judge agreement with human experts (Cohen's $\kappa$). Results in \textbf{bold} indicate cases in which a reference was better than the others. Results in {\R red} indicate cases in which the model was worse when it was ruled ``Correct'' or ``Incorrect.'' %
        }

    \label{tab:right-wrong-table}
\end{table*}

%% file: figures/affinity_bias.tex
\begin{figure*}[t!]
    \centering
    \includegraphics[width=\linewidth]{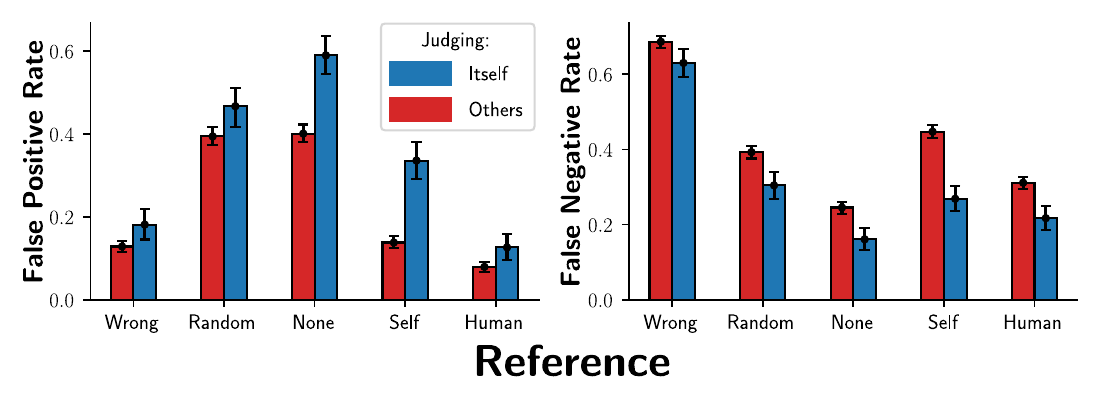}
    \caption{Error analysis for the judge models conditioned on the candidate model being graded. All models are included both as candidates and judges. \textit{Itself} (blue) refers to cases in which the judge model graded responses from itself (as a candidate). \textit{Others} (red) refers to cases in which the judge model graded responses from other models.
}
    \label{fig:affinity_bias}
\end{figure*}

%% file: sections/appendix.tex
\clearpage
\label{sec:appendix}

\section{Annotator Details}
\label{sec:annotator_details}
All of our datasets relied on two main pools of annotators: domain experts and the authors. The domain expert pool was responsible for:
\begin{enumerate}
    \item The creation of the original set of BFF-Bench questions.
    \item Writing the reference responses for BFF-Bench questions.
    \item Verifying the reference responses for BFF-Bench questions.
    \item Grading the model outputs for the BFF-Bench questions.
    \item Grading the model outputs for the (C)MT-Bench questions.
\end{enumerate}

The authors:
\begin{enumerate}
    \item Edited responses for readability and formatting.
    \item Created and verified responses for the (C)MT-Bench dataset.
    \item Graded the model outputs for the (C)MT-Bench questions.
\end{enumerate}

Our domain expert pool consisted of 9 salaried analysts employed full-time at a financial services company. Each had at least a bachelor's degree in a business or finance discipline. They were based in the United States and Pakistan. They were not compensated beyond their salary for any work performed for the creation of this dataset.

This work did not present any meaningful risk for our annotators. They were volunteers who accepted the work as part of their normal full-time duties. We were in frequent contact to ensure that they understood the goal of the work and its potential impacts. 

\subsection{BFF-Bench}
The annotation process occurred in two main stages: dataset creation and model response grading. To create the dataset we first worked with the annotators to develop a set of initial questions and draft responses. We filtered these responses to find questions that were unambiguous, difficult, and self-contained.

Once we had identified a set of questions to use as the base, we then verified the responses. A subset of the annotators were tasked with reviewing each response for accuracy. If a potential error was found, the annotator would propose a new potential response. This proposed response would then go through the same verification process.

We continued this process until we reached 80 two-turn questions with verified answers.

\subsection{(C)MT-Bench}

(C)MT-Bench was created after our initial analysis into the references in the MT-Bench dataset. We started by manually going through each provided reference in the math and reasoning category and marking them correct or incorrect. We (the authors) then fixed each error by hand, resulting in corrected reference responses.

\subsection{Grading}

The grading process for both datasets occurred within the same user interface. The annotators were given the reference response to the question as a guide and were asked to select whether the model's response was ``Correct'', ``Incorrect'', or ``Not sure''. There was an additional ``Notes'' free response box that they could use to flag any issues that arose. Because the references had been verified, we recommended that the annotators generally err on the side of trusting them. 

The exact guidelines provided to the annotators are provided in the following section (\cref{sec:annotatorguidelines}).

\subsection{Annotator Guidelines}
\label{sec:annotatorguidelines}
The model will attempt to answer the provided question (first text box above the
other two boxes).  A model will generally attempt to fool the judge with
confident wrong answers in cases where it does not know the correct
answer. Annotators will utilize the Reference (bottom left text box), which
contains the answer to the posed question (top text box).

Model Answers should include the relevant facts/formulas needed to answer the
question, with a step-by-step explanation of any mathematical manipulations.

It is important to remember that everything should be graded only with respect
to the reference answer. If you find any errors within the reference, please
mark the error in the notes! However, you should always enter the grade as if
the reference is correct.

A general set of guidelines is:
\begin{enumerate}

        \item If the model’s final answer or recommendation disagrees with the reference, it should be marked “Incorrect”.
        \item If the model agrees with the reference but its intermediate steps disagree, it should also be marked “Incorrect”.
        \item If the model agrees with the reference, all of the intermediate steps agree, and it provides the additional information that is not obviously wrong based on the reference, it should be marked “Correct”
        \item If the model agrees with the reference, all of the intermediate steps agree, but the answer is slightly off due to rounding or performing approximate calculations, it should be marked as “Correct”
        \item If the model agrees with the reference, all of the intermediate steps agree, but it is lacking some amount of information given within the reference, you should make a judgment call as to whether the missing information is crucial to the answer. If you are unsure, lean towards “Correct” or mark it “Unsure”

\end{enumerate}

\subsection{F1 Results}
\label{ref:f1_table}
\input{tables/right_wrong_table_f1_v1}

\subsection{Prompt Sensitivity}
\label{sec:prompt_sensitivity} 

In this section, we compare our main results across two prompts: the slightly modified version of the MT-Bench prompts, and the additional set of prompts we had written from scratch for this analysis. The results are displayed in \cref{tab:right-wrong-table-kappa-v1}. In general, we find the results to be very similar, demonstrating that our analysis is robust to the underlying prompt. 

\input{tables/right_wrong_double_kappa_v1}

\section{(C)MT-Bench Examples}
\label{sec:cmt_examples}

In this section, we provide an example of an error from MT-Bench.

\input{figures/sample_mt_bench_error}

\section{Confidence Intervals on Agreement}
\label{sec:confidenceintervals}

To evaluate the agreement between the judges and the human annotators' consensus, we use the Cohen's $\kappa$ metric \citep{CohenKappa}. This is an agreement measure for categorical data labeled by two raters that accounts for the chance of random agreement. It is defined as:
\begin{equation}
\kappa = 1 - \frac{1 - p_o}{1 - p_e},
\end{equation}
where $p_o$ is the observed agreement and $p_e$ is the expected chance agreement. $\kappa$ is bounded between -1 and 1, with -1 corresponding to total disagreement, 0 corresponding to random (chance) agreement, and 1 corresponding to total agreement. 

For simplicity, we used a bootstrap to compute confidence intervals for each of our metrics. Although the intervals that we compute are non-symmetric, we convert them into symmetric intervals for a more compact display in tables. To do so, we select whichever bound leads to the widest interval.

\begin{figure*}[ht]
    \centering
    \includegraphics[width=\linewidth]{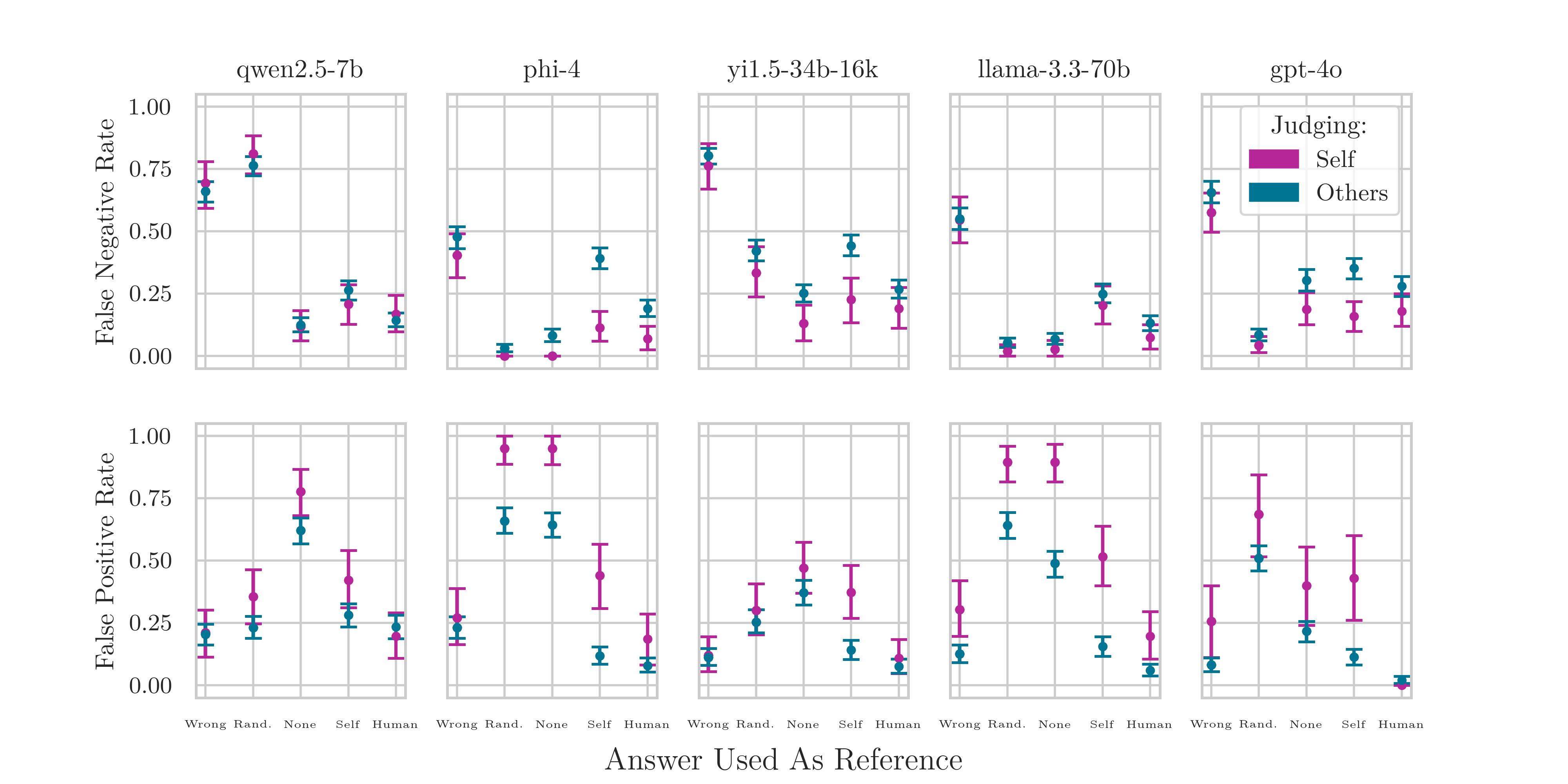}
    \caption{The full graph displaying the error rates for different LLM Judges and different reference types. The same trend displayed in \cref{fig:affinity_bias} holds for each model. Rand. refers to the Random reference type.}
    \label{fig:affinity_per_judge_all}
\end{figure*}

\section{BFF-Bench Examples}
\label{sec:bff-bench-examples}
\input{sample_bffbench_untruncated}
\input{sample_bffbench_untruncated_2}

\section{Baselines}
\label{sec:baselines}

Results for baseline methods are presented in \cref{tab:baseline_perf}. Generally we find that reward models outperform embedding models, which in turn outperform the lexical overlap methods.

\begin{table*}[ht]
\centering
\begin{tabular}{lc}
\toprule
 & Cohen's $\kappa$ \\
 \midrule
\textit{Reward Models} & \\
\texttt{INF-ORM-Llama3.1-70B} & $0.44_{\pm 0.07 }$ \\
\texttt{Llama-3.1-Nemotron-70B-Reward} & $0.37_{\pm 0.07 }$ \\
\texttt{LDL-Reward-Gemma-2-27B-v0.1} & $0.31_{\pm 0.08 }$ \\
\texttt{Skywork-Reward-Gemma-2-27B-v0.2} & $0.30_{\pm 0.07 }$ \\
\texttt{QRM-Gemma-2-27B} & $0.24_{\pm 0.08 }$ \\
\midrule
\textit{Embedding Models} & \\
\texttt{Linq-AI-Research/Linq-Embed-Mistral} & $0.35_{\pm 0.08 }$ \\
\texttt{NovaSearch/jasper\_en\_vision\_language\_v1} & $0.34_{\pm 0.07 }$ \\
\texttt{Salesforce/SFR-Embedding-Mistral} & $0.25_{\pm 0.08 }$ \\
\texttt{Alibaba-NLP/gte-Qwen2-7B-instruct} & $0.21_{\pm 0.08 }$ \\
\texttt{intfloat/e5-mistral-7b-instruct} & $0.18_{\pm 0.08 }$ \\
\midrule
\textit{{Lexical}} & \\
METEOR & $0.28_{\pm 0.08 }$ \\
ROUGE & $0.26_{\pm 0.08 }$ \\
BLEU & $0.15_{\pm 0.08 }$ \\
\bottomrule
\end{tabular}
\caption{Full results table for all baselines on the pairwise grading task.}
\label{tab:baseline_perf}
\end{table*}

\subsection{Reward Models}
We selected the top five performing models on RewardBench \citep{RewardBench}: 

\begin{itemize}
    \item \texttt{INF-ORM-Llama3.1-70B} \citep{INF-ORM-Llama3.1-70B}
    \item \texttt{LDL-Reward-Gemma-2-27B-v0.1} \citep{LDL-Reward-Gemma-2-27B-v0.1}
    \item \texttt{QRM-Gemma-2-27B} \citep{dorka2024quantile}
    \item \texttt{Skywork-Reward-Gemma-2-27B-v0.2} \citep{liu2024skywork}
    \item \texttt{Llama-3.1-Nemotron-70B-Reward} \citep{wang2024helpsteer2preferencecomplementingratingspreferences,wang2024helpsteer2}
\end{itemize}

To run the pairwise evaluation, we extracted the reward for each of the potential responses and ruled the ``more correct'' one to be the one with the higher reward.

\subsection{Reference-Based Lexical Overlap Classifiers}
We implement a series of reference based classifiers. For each of these classifiers, we compute a similarity score between the human-written reference and each of the two provided responses. We then rule the response with the higher similarity with the human-written reference to be ``Correct''.

We used the \texttt{evaluate} library (0.4.3) \citep{evaluate} to compute ROUGE, BLEU and METEOR scores. We used the default settings for each metric. 

\subsection{Embedding Models}

We selected the following five Embedding Models using \cite{mteb}:
\begin{itemize}
    \item \texttt{Linq-AI-Research/Linq-Embed-Mistral} \citep{linq}
    \item \texttt{NovaSearch/jasper\_en\_vision\_language\_v1} \citep{zhang2024jasper}
    \item \texttt{Salesforce/SFR-Embedding-Mistral} \citep{SFRAIResearch2024}
    \item \texttt{Alibaba-NLP/gte-Qwen2-7B-instruct} \citep{li2023towards}
    \item \texttt{intfloat/e5-mistral-7b-instruct} \citep{wang2023improving,wang2022text}
\end{itemize}
For the embedding metrics, we computed the cosine similarity between the embeddings of the human-written reference and the two responses. The response with the higher similarity to the reference was selected as correct.

\section{Judge Overruling}
\label{sec:judge-refusal}

The following example is \texttt{GPT-4o} grading a response from \texttt{Phi-4} using a wrong reference which inverts the definition of annuities and IRAs. The text generated by \texttt{GPT-4o} notes the inconsistencies between the reference and the response, but overrules the reference and still marks the response as correct. We observed this behavior multiple times, only with \texttt{GPT-4o}.
\\\\
\noindent\textbf{Judgment:}

The assistant's answer provides a breakdown of the differences between annuities and IRAs, similar to the reference answer. Let's compare the key elements:

1. \textit{Annuity}:
   - The assistant correctly identifies annuities as financial products designed to provide a steady income stream, and they can be immediate or deferred. It mentions different types of annuities (fixed, variable, indexed) and notes the tax-deferred nature of annuities and the penalties for early withdrawal.
   - The reference answer describes annuities as investment accounts for holding financial assets, emphasizing accumulating contributions and interest until retirement age, and mentions custodial fees and mutual funds' expense ratios.
   - There is a discrepancy in the definition: the reference answer incorrectly describes annuities as accounts for holding investments, which aligns more with the description of IRAs.

2. \textit{IRA}:
   - The assistant correctly describes IRAs as tax-advantaged savings accounts with different types (Traditional and Roth) and their unique tax implications. It covers tax treatment, access to funds, and ownership.
   - The reference answer incorrectly labels IRAs as insurance products, which is inaccurate. It also incorrectly suggests IRAs do not have contribution limits, which is a mistake.

3. \textit{Key Differences}:
   - The assistant highlights investment control, flexibility, tax treatment, and income stream as key differences, which are generally accurate and align with the correct understanding of annuities and IRAs.

Overall, the reference answer contains significant inaccuracies regarding the fundamental nature of annuities and IRAs. In contrast, the assistant's answer provides a more accurate description and comparison of these financial products.

[[Correct]]

\section{LLM Judge Ablations}
\label{sec:judge-ablation}

We ablate two major design decisions for our LLM Judges. We use the GPT-4o judge for each ablation. \cref{tab:temperature-ablation-table} evaluates the impact of temperature and \cref{tab:temp-cot-ablation-table} evaluates the influence of chain-of-thought. We do not observe a sensitivity to either of these two factors, providing evidence that our results are robust across these two hyperparameters. 

\begin{table}[t!]
    \centering
    \begin{tabular}{lc|ccc}
        \toprule
         \textbf{Ref} & \textbf{Stratum} & $\mathbf{T = 0.0}$ & $\mathbf{T = 0.2}$ & $\mathbf{T = 0.7}$ \\
        \midrule
\multirow{2}{*}{None} & Correct & $0.39_{\pm 0.07}$ & $0.45_{\pm 0.07}$ & $0.42_{\pm 0.07}$\\
& Incorrect & $0.25_{\pm 0.15}$ & $0.16_{\pm 0.14}$ & $0.15_{\pm 0.14}$\\
\midrule
\multirow{2}{*}{Self} & Correct & $0.40_{\pm 0.06}$ & $0.40_{\pm 0.06}$ & $0.49_{\pm 0.06}$\\
& Incorrect & $0.21_{\pm 0.16}$ & $0.22_{\pm 0.16}$ & $0.13_{\pm 0.14}$\\
\midrule
\multirow{2}{*}{Human} & Correct & $0.49_{\pm 0.06}$ & $0.49_{\pm 0.05}$ & $0.51_{\pm 0.05}$\\
& Incorrect & $0.72_{\pm 0.17}$ & $0.72_{\pm 0.16}$ & $0.69_{\pm 0.17}$\\
        \bottomrule
    \end{tabular}
    \caption{Ablation of temperature for the GPT-4o Judge. The LLM Judge performance is not sensitive to the temperature for this model.}
    \label{tab:temperature-ablation-table}
\end{table}

\begin{table}[t!]
    \centering
    \begin{tabular}{l|ccc}
        \toprule
         \textbf{Config} & \textit{None} & \textit{Self} & \textit{Human} \\
        \midrule
T=0.0 (CoT) & $0.44_{\pm 0.06}$ & $0.44_{\pm 0.05}$ & $0.59_{\pm 0.05}$\\
T=0.2 (CoT) & $0.46_{\pm 0.06}$ & $0.43_{\pm 0.06}$ & $0.59_{\pm 0.05}$\\
T=0.7 (CoT) & $0.44_{\pm 0.06}$ & $0.50_{\pm 0.06}$ & $0.60_{\pm 0.05}$\\
T=0.7 (No-CoT) & $0.46_{\pm 0.06}$ & $0.47_{\pm 0.06}$ & $0.59_{\pm 0.05}$\\
        \bottomrule
    \end{tabular}
    \caption{Ablation of chain-of-thought for the GPT-4o Judge. We do not observe significant shifts in performance with the addition of chain-of-thought.}
    \label{tab:temp-cot-ablation-table}
\end{table}

\section{Jury Experiment}
\label{sec:jury}

We constructed a majority-vote Jury using all five judges. To see if our main results hold in the jury setting, we split the questions based on whether the majority of the jury (at least 3 of the 5) got the question correct or incorrect using the same expert labels used in the rest of the analysis. The results from this experiment can be viewed in \cref{tab:correct-incorrect-counts-table}.

\begin{table*}[t!]
    \centering
    \resizebox{\textwidth}{!}{

    \begin{tabular}{l|ccc|ccc}
        \toprule
       & \multicolumn{3}{c|}{\textbf{Pairwise}} & \multicolumn{3}{c}{\textbf{Single}} \\
         \textbf{Stratum} & \textit{None} & \textit{Self} & \textit{Human} & \textit{None} & \textit{Self} & \textit{Human} \\
        \midrule
Majority Correct & $0.51_{\pm 0.07}$ & $0.75_{\pm 0.07}$  & $0.81_{\pm 0.05}$  & $0.36_{\pm 0.10}$  & $0.51_{\pm 0.08}$ & $0.58_{\pm 0.08}$ \\
Majority Incorrect & $0.40_{\pm 0.11}$ & $0.46_{\pm 0.14}$  & $0.85_{\pm 0.09}$  & $0.07_{\pm 0.06}$ & $0.01_{\pm 0.12}$  & $0.51_{\pm 0.16}$ \\
        \bottomrule
    \end{tabular}
    }
    \caption{Majority Jury performance using all five LLM Judges, conditioned on whether the majority were marked as correct or incorrect.}
    \label{tab:correct-incorrect-counts-table}
\end{table*}

The jury does not escape the core limitation we identified. These results additionally offer an alternative explanation as to why a jury might aid judgment performance. By aggregating based on the majority, the jury in effect can reduce the overall number of items for which it is collectively incorrect. Performance on those questions, however, still remains poor (e.g., near 0 for the Single case) and is recovered only by providing a human reference. 

\section{Domain Generalization}
\label{sec:domain-generalization}
\input{tables/domain_generalization}

To demonstrate that our main results are not linked to any one domain, we split agreement metrics over our two main datasets in \cref{tab:correctness-conditioned-kappa}. We aggregated over each LLM Judge, and computed the confidence intervals of the change in agreement when incorrect for each reference type. $\kappa_{\text{Correct}}$ is the Cohen’s $\kappa$ on questions the Judge got correct, $\kappa_{\text{Incorrect}}$ is the Cohen’s $\kappa$ on questions it got incorrect, and $\Delta\kappa$ is the average change in $\kappa$ when the Judge is correct (e.g. +0.249 means the LLM Judge has a $\kappa$ 0.249 higher when it was judged to have answered the question correctly than when it was not). We include the 95\% CI for $\Delta\kappa$ as well.

Across both datasets, when the Judge is not provided a reference (None, Self) there is a large difference in $\kappa$. When the model is provided a reference, not only does the overall agreement increase but the $\Delta\kappa$ decreases. These results demonstrate that the effect is not specific to either of the two source datasets.

\section{Expanded Models}
\label{sec:more-models}

We selected our initial set of models to try to emphasize diversity of both size and family rather than recency. In \cref{tab:right-wrong-table-frontier}, we present extended results on a set of frontier models. We used the \texttt{GPT-4o} Judge with the human-verified references as a proxy correctness label. Although this is only a proxy for our human experts, we did verify in \cref{fig:overall-performance} that this LLM Judge and reference combination reaches high agreement with our human annotators. Because of this, we regard these results as corroboration rather than equivalent to our main table.

\input{tables/correct_incorrect_frontier}

Additionally, we ablate the reasoning level for these models in \cref{tab:reasoning-ablation-table} but find little overall impact. Notably, higher reasoning effort can narrow the no-reference gap (e.g., \texttt{GPT-5.4} at high effort), which is consistent with our central finding: as the judge becomes better able to answer the question itself, its reliance on a reference decreases. The gap does not close entirely, and a human reference still yields the most reliable judgments.

\input{tables/reasoning_ab}

\newpage
\section{Judgment Prompts}
\label{sec:judgeprompts}
\input{judge_prompts}

%% file: tables/right_wrong_table_f1_v1.tex
\begin{table*}[ht!]

    \resizebox{\textwidth}{!}{ %
    \begin{tabular}{l|ccc|ccc|ccc}
    \toprule
    & \multicolumn{9}{c}{\bf Single Grading} \\
    \hline
    & \multicolumn{3}{c|}{Overall} & \multicolumn{3}{c|}{Judge Answered Correctly} & \multicolumn{3}{c}{Judge Answered Incorrectly}\\
& \textit{None} & \textit{Self} & \textit{Human} & \textit{None} & \textit{Self} & \textit{Human} & \textit{None} & \textit{Self} & 
         \textit{Human} \\
\hline
GPT-4o & $0.77_{\pm 0.03 }$ & $0.77_{\pm 0.03 }$ & $\textbf{0.85}_{\pm 0.02 }$ & $0.83_{\pm 0.03 }$ & $0.83_{\pm 0.03 }$ & $0.86_{\pm 0.02 }$ & \R $0.34_{\pm 0.13 }$ & \R $0.30_{\pm 0.13 }$ & $\textbf{0.84}_{\pm 0.12 }$ \\
Llama 3.3 70b & $0.81_{\pm 0.02 }$ & $0.80_{\pm 0.03 }$ & $\textbf{0.91}_{\pm 0.02 }$ & $0.93_{\pm 0.02 }$ & $0.88_{\pm 0.02 }$ & $0.94_{\pm 0.02 }$ & \R $0.39_{\pm 0.07 }$ & \R $0.41_{\pm 0.09 }$ & \R $\textbf{0.76}_{\pm 0.09 }$ \\
Phi 4 & $0.78_{\pm 0.02 }$ & $0.74_{\pm 0.03 }$ & $\textbf{0.88}_{\pm 0.02 }$ & $0.92_{\pm 0.02 }$ & $0.84_{\pm 0.03 }$ & $0.92_{\pm 0.02 }$ & \R $0.32_{\pm 0.07 }$ & \R $0.22_{\pm 0.10 }$ & \R $\textbf{0.65}_{\pm 0.10 }$ \\
Yi 1.5 34B & $0.76_{\pm 0.03 }$ & $0.69_{\pm 0.03 }$ & $\textbf{0.83}_{\pm 0.02 }$ & $0.87_{\pm 0.03 }$ & $0.80_{\pm 0.03 }$ & $0.88_{\pm 0.03 }$ & \R $0.46_{\pm 0.07 }$ & \R $0.37_{\pm 0.08 }$ & \R $\textbf{0.75}_{\pm 0.07 }$ \\
Qwen 2.5 7B & $0.76_{\pm 0.03 }$ & $0.76_{\pm 0.03 }$ & $\textbf{0.85}_{\pm 0.02 }$ & $0.89_{\pm 0.02 }$ & $0.86_{\pm 0.03 }$ & $0.91_{\pm 0.02 }$ & \R $0.45_{\pm 0.07 }$ & \R $0.45_{\pm 0.07 }$ & \R $\textbf{0.70}_{\pm 0.06 }$ \\

\end{tabular}

}
\vspace{1em}
    \resizebox{\textwidth}{!}{ %
    \begin{tabular}{l|ccc|ccc|ccc}
    \toprule
    & \multicolumn{9}{c}{\bf Pairwise Grading} \\
    \hline
    & \multicolumn{3}{c|}{Overall} & \multicolumn{3}{c|}{Judge Answered Correctly} & \multicolumn{3}{c}{Judge Answered Incorrectly}\\
         & \textit{None} & \textit{Self} & \textit{Human} & \textit{None} & \textit{Self} & \textit{Human} & \textit{None} & \textit{Self} & 
         \textit{Human} \\

\hline
GPT-4o & $0.82_{\pm 0.02 }$ & $0.86_{\pm 0.01 }$ & $\textbf{0.95}_{\pm 0.01 }$ & $0.88_{\pm 0.01 }$ & $0.94_{\pm 0.01 }$ & $0.95_{\pm 0.01 }$ & \R $0.62_{\pm 0.06 }$ & \R $0.54_{\pm 0.05 }$ & \R $\textbf{0.90}_{\pm 0.03 }$ \\
Llama 3.3 70b & $0.82_{\pm 0.02 }$ & $0.85_{\pm 0.02 }$ & $\textbf{0.96}_{\pm 0.01 }$ & $0.89_{\pm 0.02 }$ & $0.93_{\pm 0.01 }$ & $\textbf{0.96}_{\pm 0.01 }$ & \R $0.66_{\pm 0.04 }$ & \R $0.65_{\pm 0.04 }$ & $\textbf{0.95}_{\pm 0.02 }$ \\
Phi 4 & $0.75_{\pm 0.02 }$ & $0.80_{\pm 0.02 }$ & $\textbf{0.94}_{\pm 0.01 }$ & $0.88_{\pm 0.02 }$ & $0.95_{\pm 0.01 }$ & $\textbf{0.98}_{\pm 0.01 }$ & \R $0.49_{\pm 0.06 }$ & \R $0.49_{\pm 0.06 }$ & \R $\textbf{0.89}_{\pm 0.03 }$ \\
Yi 1.5 34B & $0.42_{\pm 0.05 }$ & $0.50_{\pm 0.04 }$ & $\textbf{0.81}_{\pm 0.03 }$ & $0.62_{\pm 0.08 }$ & $0.79_{\pm 0.06 }$ & $0.86_{\pm 0.05 }$ & \R $0.27_{\pm 0.06 }$ & \R $0.33_{\pm 0.07 }$ & $\textbf{0.79}_{\pm 0.04 }$ \\
Qwen 2.5 7B & $0.67_{\pm 0.03 }$ & $0.69_{\pm 0.03 }$ & $\textbf{0.85}_{\pm 0.02 }$ & $0.81_{\pm 0.03 }$ & $0.83_{\pm 0.03 }$ & $\textbf{0.89}_{\pm 0.02 }$ & \R $0.47_{\pm 0.05 }$ & \R $0.49_{\pm 0.05 }$ & \R $\textbf{0.78}_{\pm 0.04 }$ \\

\bottomrule
\end{tabular}
}
\caption{Overall results using F1 instead of Cohen's $\kappa$. Cells in \colorbox{red!30}{red} represent cases in which the model is worse when it was wrong as a candidate ($\alpha < 0.05$). Numbers in \textbf{bold} indicate the maximum values on that subset with statistical significance ($\alpha < 0.05$).}
\label{tab:right-wrong-table-f1-v1}
\end{table*}

%% file: tables/right_wrong_double_kappa_v1.tex
\begin{table*}[ht!]

    \resizebox{\textwidth}{!}{ %
    \begin{tabular}{l|ccc|ccc|ccc}
    \toprule
    & \multicolumn{9}{c}{\bf Prompt One} \\
    \hline
    & \multicolumn{3}{c|}{Overall} & \multicolumn{3}{c|}{Judge Answered Correctly} & \multicolumn{3}{c}{Judge Answered Incorrectly}\\
& \textit{None} & \textit{Self} & \textit{Human} & \textit{None} & \textit{Self} & \textit{Human} & \textit{None} & \textit{Self} & 
         \textit{Human} \\
\hline
GPT-4o & $0.47_{\pm 0.05 }$ & $0.52_{\pm 0.05 }$ & $\textbf{0.68}_{\pm 0.04 }$ & $0.46_{\pm 0.07 }$ & $0.52_{\pm 0.06 }$ & \R $0.59_{\pm 0.06 }$ & \R $0.16_{\pm 0.14 }$ & \R $0.13_{\pm 0.15 }$ & $\textbf{0.81}_{\pm 0.13 }$ \\
Llama 3.3 70b & $0.42_{\pm 0.05 }$ & $0.53_{\pm 0.05 }$ & $\textbf{0.78}_{\pm 0.04 }$ & $0.62_{\pm 0.09 }$ & $0.54_{\pm 0.08 }$ & $0.74_{\pm 0.07 }$ & \R $0.13_{\pm 0.06 }$ & \R $0.23_{\pm 0.11 }$ & $\textbf{0.69}_{\pm 0.10 }$ \\
Phi 4 & $0.27_{\pm 0.05 }$ & $0.47_{\pm 0.05 }$ & $\textbf{0.72}_{\pm 0.04 }$ & $0.48_{\pm 0.10 }$ & $0.47_{\pm 0.07 }$ & $\textbf{0.67}_{\pm 0.07 }$ & \R $0.03_{\pm 0.05 }$ & \R $0.03_{\pm 0.11 }$ & $\textbf{0.56}_{\pm 0.12 }$ \\
Yi 1.5 34B & $0.38_{\pm 0.06 }$ & $0.37_{\pm 0.05 }$ & $\textbf{0.63}_{\pm 0.05 }$ & $0.23_{\pm 0.11 }$ & $0.29_{\pm 0.07 }$ & \R $0.43_{\pm 0.09 }$ & $0.18_{\pm 0.10 }$ & $0.14_{\pm 0.10 }$ & $\textbf{0.65}_{\pm 0.08 }$ \\
Qwen 2.5 7B & $0.24_{\pm 0.06 }$ & $0.43_{\pm 0.06 }$ & $\textbf{0.63}_{\pm 0.05 }$ & $0.31_{\pm 0.10 }$ & $0.37_{\pm 0.10 }$ & $0.54_{\pm 0.09 }$ & \R $0.08_{\pm 0.07 }$ & $0.21_{\pm 0.10 }$ & $\textbf{0.57}_{\pm 0.08 }$ \\

\end{tabular}

}
\vspace{1em}
    \resizebox{\textwidth}{!}{ %
    \begin{tabular}{l|ccc|ccc|ccc}
    \toprule
    & \multicolumn{9}{c}{\bf Prompt Two} \\
    \hline
    & \multicolumn{3}{c|}{Overall} & \multicolumn{3}{c|}{Judge Answered Correctly} & \multicolumn{3}{c}{Judge Answered Incorrectly}\\
         & \textit{None} & \textit{Self} & \textit{Human} & \textit{None} & \textit{Self} & \textit{Human} & \textit{None} & \textit{Self} & 
         \textit{Human} \\

\hline
GPT-4o & $0.47_{\pm 0.06 }$ & $0.53_{\pm 0.06 }$ & $\textbf{0.65}_{\pm 0.05 }$ & $0.47_{\pm 0.07 }$ & $0.53_{\pm 0.07 }$ & $0.57_{\pm 0.07 }$ & \R $0.10_{\pm 0.14 }$ & \R $0.10_{\pm 0.16 }$ & $\textbf{0.67}_{\pm 0.21 }$ \\
Llama 3.3 70b & $0.38_{\pm 0.06 }$ & $0.55_{\pm 0.06 }$ & $\textbf{0.76}_{\pm 0.04 }$ & $0.56_{\pm 0.10 }$ & $0.58_{\pm 0.08 }$ & $0.71_{\pm 0.07 }$ & \R $0.11_{\pm 0.05 }$ & \R $0.21_{\pm 0.11 }$ & $\textbf{0.69}_{\pm 0.09 }$ \\
Phi 4 & $0.33_{\pm 0.06 }$ & $0.55_{\pm 0.05 }$ & $\textbf{0.69}_{\pm 0.05 }$ & $0.45_{\pm 0.09 }$ & $0.61_{\pm 0.07 }$ & $0.71_{\pm 0.07 }$ & \R $0.08_{\pm 0.06 }$ & \R $0.06_{\pm 0.11 }$ & \R $\textbf{0.43}_{\pm 0.11 }$ \\
Yi 1.5 34B & $0.22_{\pm 0.06 }$ & $0.40_{\pm 0.06 }$ & $\textbf{0.63}_{\pm 0.05 }$ & $0.25_{\pm 0.13 }$ & $0.35_{\pm 0.11 }$ & $\textbf{0.62}_{\pm 0.10 }$ & \R $0.05_{\pm 0.06 }$ & $0.17_{\pm 0.09 }$ & $\textbf{0.50}_{\pm 0.08 }$ \\
Qwen 2.5 7B & $0.29_{\pm 0.06 }$ & $0.46_{\pm 0.06 }$ & $\textbf{0.63}_{\pm 0.05 }$ & $0.32_{\pm 0.10 }$ & $0.37_{\pm 0.09 }$ & $0.50_{\pm 0.09 }$ & \R $0.10_{\pm 0.08 }$ & $0.23_{\pm 0.11 }$ & $\textbf{0.59}_{\pm 0.09 }$ \\

\bottomrule
\end{tabular}
}
\caption{Performance comparison on single grading using the two sets of prompts. Cells in \colorbox{red!30}{red} represent cases in which the model is worse when it was wrong as a candidate ($\alpha < 0.05$). Numbers in \textbf{bold} indicate the maximum values on that subset with statistical significance ($\alpha < 0.05$).}
\label{tab:right-wrong-table-kappa-v1}
\end{table*}

%% file: figures/sample_mt_bench_error.tex
\begin{figure}[ht!]
\small
\begin{tcolorbox}[
        enhanced jigsaw, 
        colback=gray!5!white, 
        colframe=gray!90!black, 
        boxrule=1pt, 
        arc=5pt, 
        sharp corners=downhill, 
        left=0pt, right=0pt, top=0pt, bottom=0pt,
        title=MT-Bench Question
    ]
  The vertices of a triangle are at points (0, 0), (-1, 1), and (3, 3). What is the area of the triangle?
\end{tcolorbox}

 \begin{tcolorbox}[
        enhanced jigsaw, 
        colback=providedbg, 
        colframe=providedborder, 
        boxrule=1pt, 
        arc=5pt, 
        sharp corners=downhill, 
        left=0pt, right=0pt, top=0pt, bottom=0pt,
        title=MT-Bench Reference
    ]
To find the area of the triangle, we can use the formula for the area of a triangle with coordinates A(x1, y1), B(x2, y2), and C(x3, y3):

\vspace{0.25em}

Area = (1/2) * |(x1(y2 - y3) + x2(y3 - y1) + x3(y1 - y2))|

\vspace{0.25em}

In this case, the coordinates are A(0, 0), B(-1, 1), and C(3, 3). Plugging these values into the formula, we get:

\vspace{0.25em}

Area = (1/2) * |(0(1 - 3) + (-1)(3 - 0) + 3(0 - 1))|

Area = (1/2) * |(-0 + \hl{{\bf 3}} - 3)|

Area = (1/2) * |0|

Area = 0

\vspace{0.25em}

The area of the triangle is 0. This means that the three points are collinear, and there is no triangle formed by these points.
\end{tcolorbox}

\caption{Example of an incorrect reference included in MT-Bench (\hl{\bf 3} should be $-3$). The correct answer is that the area of the triangle is 3.}
\label{fig:mt-bench-wrong}
\end{figure}

%% file: sample_bffbench_untruncated.tex
\begin{figure}[ht!]

\centering
\small

\begin{tcolorbox}[
    boxsep=1pt,
    left=2pt,right=2pt,top=2pt,bottom=2pt,
    colback=blue!5!white,
    colframe=blue!40!black,
    before skip=2pt,
    after skip=2pt,
    sharp corners
]
\textbf{Question (Turn One):}
What is the IFRS 16 treatment of short term or low value leased assets? 
What are the restrictions?
\end{tcolorbox}

\begin{tcolorbox}[
    boxsep=1pt,
    left=2pt,right=2pt,top=2pt,bottom=2pt,
    colback=green!2!white,
    colframe=green!25!black,
    before skip=1pt,
    after skip=3pt,
    sharp corners
]
\small
\textbf{Gold Response:}
IFRS 16 Leases permit a simplified treatment for assets with a lease period of 12 months or less, or of low value. Although the standard does not give a numerical definition of 'low value' it does give examples of the types of assets that may be included, and this includes telephones. The simplified treatment allows the lease payments to be charged as an expense over the lease period, applying the accruals concept.

IFRS 16 Leases does not specify a particular monetary amount below which an asset would be considered 'low value', although the basis for conclusion indicates a value of \$5,000 as a guide. The standard also gives the following examples of low-value assets:

- tablets

- small personal computers

- telephones

- small items of furniture.

The assessment of whether an asset qualifies as having a 'low value' must be made based on its value when new. Therefore, a car would not qualify as a low-value asset, even if it was very old at the commencement of the lease.
\end{tcolorbox}

\begin{tcolorbox}[
    boxsep=1pt,
    left=2pt,right=2pt,top=2pt,bottom=2pt,
    colback=blue!5!white,
    colframe=blue!40!black,
    before skip=3pt,
    after skip=1pt,
    sharp corners
]
\small
\textbf{Question (Turn Two):}
On April 1st 2023, Abigail acquired telephones for her salesforce under a 
two-year lease agreement. The terms of the lease require an initial payment 
of \$3,000, followed by a payment of \$9,000 on March 31st 2024 and an 
additional \$9,000 on March 31st 2025. Show the impact of this lease 
arrangement on Abigail's financial statements for the year ended 
31 December 2023 under IFRS 16?
\end{tcolorbox}

\begin{tcolorbox}[
    boxsep=1pt,
    left=2pt,right=2pt,top=2pt,bottom=2pt,
    colback=green!2!white,
    colframe=green!25!black,
    before skip=1pt,
    after skip=2pt,
    sharp corners=downhill,
]
\small
\textbf{Gold Response:}
\[
\begin{aligned}
\text{Annual lease rental expense} &= \frac{\text{Total rentals payable}}{\text{Total lease period}} = \frac{\$3,000 + \$9,000 + \$9,000}{2\text{ years}} \\[6pt]
&= \$10,500\text{ per annum} \\[8pt]
\text{Expense to 31 December 2023} &= \$10,500 \times \frac{9}{12} = \$7,875 \\[8pt]
\text{Accrued expense} &= \$7,875 - \$3,000 = \$4,875
\end{aligned}
\]

The expense in this period of \$7,875 is not the same as the payment of \$3,000, so we need to accrue an additional expense of \$4,875.
\end{tcolorbox}

\caption{Untruncated example from BFF-Bench.}
\label{fig:bffbench-sample-full}
\end{figure}

%% file: sample_bffbench_untruncated_2.tex
\begin{figure}[ht!]
\centering
\small

\begin{tcolorbox}[
    boxsep=1pt, left=2pt,right=2pt,top=2pt,bottom=2pt,
    colback=blue!5!white, colframe=blue!40!black,
    before skip=2pt, after skip=2pt, sharp corners
]
\textbf{Question (Turn One):}  
What is the loan-to-value (LTV) ratio for Class A of the following structured finance securities? Assume a loan amount of \$407 million and an as-is appraised value of the underlying securities of \$550,750,000. Also, keep in mind that the total required credit risk retention percentage (RR interest) for this transaction is 5.00\%.

\begin{tabular}{lc}

CLASS & PRELIMINARY AMOUNT \$) \\
A & 183,160,000 \\
B & 59,185,000 \\
C & 43,890,000 \\
D & 58,178,000 \\
E & 42,237,000 \\
RR interest & 20,350,000 \\

\end{tabular}
\end{tcolorbox}

\begin{tcolorbox}[
    boxsep=1pt, left=2pt,right=2pt,top=2pt,bottom=2pt,
    colback=green!2!white, colframe=green!25!black,
    before skip=1pt, after skip=3pt, sharp corners
]
\small
\textbf{Gold Response:}
1.) The total loan amount is \$407 million.

2.) The as-is appraised value of the underlying collateral is \$550,744,248.99.

3.) The required RR interest (credit risk retention) is 5.00\% of the total loan amount, or 20,350,000.

The RR interest represents the portion of the loan that the sponsor retains in order to share the risk. This amount is subtracted from the appraised value since it's not part of the bondholder allocation.
\[
\$550,744,248 - \$20,350,000 =  \$530,394,248.99        
\]
Next, divide the Class A balance by \$530,394,248.99: 
\[
\$183,160,000 / \$530,394,248.99 = 0.3453
\]
The LTV of Class A is approximately 34.53\%, or 35\% when rounded to the nearest percent.
\end{tcolorbox}

\begin{tcolorbox}[
    boxsep=1pt, left=2pt,right=2pt,top=2pt,bottom=2pt,
    colback=blue!5!white, colframe=blue!40!black,
    before skip=3pt, after skip=1pt, sharp corners
]
\small
\textbf{Question (Turn Two):}  
If the RR interest balance of \$20,350,000 is instead allocated to Class E, how does this affect the loan-to-value (LTV) ratio for Class A?
\end{tcolorbox}

\begin{tcolorbox}[
    boxsep=1pt, left=2pt,right=2pt,top=2pt,bottom=2pt,
    colback=green!2!white, colframe=green!25!black,
    before skip=1pt, after skip=2pt, sharp corners=downhill,
]
\small
\textbf{Gold Response:}
In this scenario, the new balance of Class E is the original Class E balance of  \$42,237,000 plus the previous RR balance of \$20,350,000, or \$62,587,000.

Since the RR class is no longer a factor, the Appraisal value is \$550,750,000. Therefore, the Class A LTV is simply the Class A balance divided by the Appraisal value, or 
\[
183,160,000 / \$550,750,000 = ~33.24\%,
\]
or 33\% if rounded to the nearest percent. 

\end{tcolorbox}

\caption{Untruncated example from BFF-Bench.}
\label{fig:bffbench-sample-full-2}
\end{figure}

%% file: tables/domain_generalization.tex
\begin{table*}[t]
\centering
\small

\begin{tabular}{lcccc}
\toprule
\multicolumn{5}{c}{\textbf{Single BFF-Bench}} \\
\midrule
Reference & $\kappa_{\text{Correct}}$ & $\kappa_{\text{Incorrect}}$ & $\Delta\kappa$ & 95\% CI \\
\midrule
None  & 0.333 & 0.085 & +0.249 & [+0.19, +0.31] \\
Self  & 0.430 & 0.160 & +0.270 & [+0.20, +0.34] \\
Human & 0.512 & 0.531 & $-0.019$ & [$-0.08$, +0.05] \\
\bottomrule
\end{tabular}

\vspace{1.2em}

\begin{tabular}{lcccc}
\toprule
\multicolumn{5}{c}{\textbf{Single (C)MT-Bench}} \\
\midrule
Reference & $\kappa_{\text{Correct}}$ & $\kappa_{\text{Incorrect}}$ & $\Delta\kappa$ & 95\% CI \\
\midrule
None  & 0.469 & 0.207 & +0.263 & [+0.11, +0.42] \\
Self  & 0.607 & 0.112 & +0.495 & [+0.36, +0.64] \\
Human & 0.753 & 0.640 & +0.112 & [$-0.00$, +0.23] \\
\bottomrule
\end{tabular}

\vspace{1.2em}

\begin{tabular}{lcccc}
\toprule
\multicolumn{5}{c}{\textbf{Pairwise BFF-Bench}} \\
\midrule
Reference & $\kappa_{\text{Correct}}$ & $\kappa_{\text{Incorrect}}$ & $\Delta\kappa$ & 95\% CI \\
\midrule
None  & 0.409 & 0.183 & +0.226 & [+0.17, +0.28] \\
Self  & 0.638 & 0.139 & +0.499 & [+0.44, +0.56] \\
Human & 0.747 & 0.587 & +0.160 & [+0.10, +0.22] \\
\bottomrule
\end{tabular}

\vspace{1.2em}

\begin{tabular}{lcccc}
\toprule
\multicolumn{5}{c}{\textbf{Pairwise (C)MT-Bench}} \\
\midrule
Reference & $\kappa_{\text{Correct}}$ & $\kappa_{\text{Incorrect}}$ & $\Delta\kappa$ & 95\% CI \\
\midrule
None  & 0.464 & 0.256 & +0.208 & [+0.10, +0.31] \\
Self  & 0.708 & 0.244 & +0.465 & [+0.35, +0.58] \\
Human & 0.880 & 0.784 & +0.096 & [+0.01, +0.18] \\
\bottomrule
\end{tabular}
\caption{Agreement between the LLM judges and human annotators, conditioned on whether the judge answered the question correctly. $\kappa_{\text{Correct}}$ and $\kappa_{\text{Incorrect}}$ denote Cohen's $\kappa$ computed on the subsets of questions where the judge was respectively correct or incorrect. $\Delta\kappa = \kappa_{\text{Correct}} - \kappa_{\text{Incorrect}}$. Confidence intervals are computed via bootstrap.}
\label{tab:correctness-conditioned-kappa}

\end{table*}

%% file: tables/correct_incorrect_frontier.tex
\begin{table*}[t!]
    \centering
    \resizebox{\textwidth}{!}{

    \begin{tabular}{lc|ccc|ccc}
        \toprule
       & \textbf{As Candidate,} & \multicolumn{3}{c|}{\textbf{Pairwise Grading}} & \multicolumn{3}{c}{\textbf{Single Grading}} \\
         \textbf{Judge}  & \textbf{Judge was:} & \textit{None} & \textit{Self} & \textit{Human} & \textit{None} & \textit{Self} & \textit{Human} \\
        \midrule
\multirow{2}{*}{Opus 4.7} & Correct & $0.86_{\pm 0.04}$ & $0.90_{\pm 0.04}$ & $0.91_{\pm 0.04}$ & $0.63_{\pm 0.06}$ & $0.69_{\pm 0.06}$ & $0.77_{\pm 0.05}$\\
& Incorrect & $0.33_{\pm 0.16}$ & $0.33_{\pm 0.15}$ & $0.66_{\pm 0.13}$ & $0.26_{\pm 0.14}$ & $0.27_{\pm 0.12}$ & $0.62_{\pm 0.11}$\\
\midrule
\multirow{2}{*}{GPT-5.4} & Correct & $0.75_{\pm 0.06}$ & $0.87_{\pm 0.04}$ & $0.89_{\pm 0.04}$ & $0.51_{\pm 0.06}$ & $0.64_{\pm 0.06}$ & $0.66_{\pm 0.06}$\\
& Incorrect & $0.46_{\pm 0.11}$ & $0.45_{\pm 0.13}$ & $0.87_{\pm 0.08}$ & $0.32_{\pm 0.11}$ & $0.25_{\pm 0.12}$ & $0.67_{\pm 0.09}$\\
\midrule
\multirow{2}{*}{Gemini 3.1} & Correct & $0.78_{\pm 0.06}$ & $0.80_{\pm 0.06}$ & $0.83_{\pm 0.05}$ & $0.55_{\pm 0.06}$ & $0.62_{\pm 0.06}$ & $0.69_{\pm 0.06}$\\
& Incorrect & $0.67_{\pm 0.13}$ & $0.62_{\pm 0.14}$ & $0.95_{\pm 0.07}$ & $0.44_{\pm 0.13}$ & $0.35_{\pm 0.14}$ & $0.71_{\pm 0.10}$\\
        \bottomrule
    \end{tabular}
    }
    \caption{Recreation of \cref{tab:right-wrong-table} using an LLM Judge (GPT-4o with Human references) as a proxy correctness label on a series of frontier models. These results demonstrate the same overall behavior observed in the main experiment: high agreement on the questions in which they were judged to have answered correctly, large lift in performance for using the Human references on questions in which they were judged to be incorrect. }
    \label{tab:right-wrong-table-frontier}
\end{table*}

%% file: tables/reasoning_ab.tex
\begin{table*}[t!]
    \centering
    \resizebox{\textwidth}{!}{

    \begin{tabular}{lcc|cc|cc}
        \toprule
       & & \textbf{As Candidate,} & \multicolumn{2}{c|}{\textbf{Pairwise Grading}} & \multicolumn{2}{c}{\textbf{Single Grading}} \\
         \textbf{Judge} & \textbf{Reasoning} & \textbf{Judge was:} & \textit{None} & \textit{Human} & \textit{None} & \textit{Human} \\
        \midrule
\multirow{6}{*}{Opus 4.7} & \multirow{2}{*}{low} & Correct & $0.78_{\pm 0.06}$ & $0.90_{\pm 0.04}$ & $0.63_{\pm 0.06}$ & $0.77_{\pm 0.05}$\\
& & Incorrect & $0.39_{\pm 0.14}$ & $0.66_{\pm 0.12}$ & $0.33_{\pm 0.12}$ & $0.65_{\pm 0.10}$\\
\cmidrule{2-7}
& \multirow{2}{*}{default} & Correct & $0.86_{\pm 0.04}$ & $0.91_{\pm 0.04}$ & $0.63_{\pm 0.06}$ & $0.77_{\pm 0.05}$\\
& & Incorrect & $0.33_{\pm 0.16}$ & $0.66_{\pm 0.13}$ & $0.26_{\pm 0.14}$ & $0.62_{\pm 0.11}$\\
\cmidrule{2-7}
& \multirow{2}{*}{high} & Correct & $0.86_{\pm 0.05}$ & $0.91_{\pm 0.04}$ & $0.65_{\pm 0.06}$ & $0.77_{\pm 0.05}$\\
& & Incorrect & $0.40_{\pm 0.16}$ & $0.80_{\pm 0.11}$ & $0.27_{\pm 0.15}$ & $0.67_{\pm 0.10}$\\
\midrule
\multirow{6}{*}{GPT-5.4} & \multirow{2}{*}{low} & Correct & $0.85_{\pm 0.05}$ & $0.90_{\pm 0.04}$ & $0.54_{\pm 0.06}$ & $0.57_{\pm 0.05}$\\
& & Incorrect & $0.55_{\pm 0.13}$ & $0.95_{\pm 0.06}$ & $0.34_{\pm 0.11}$ & $0.56_{\pm 0.09}$\\
\cmidrule{2-7}
& \multirow{2}{*}{default} & Correct & $0.75_{\pm 0.06}$ & $0.89_{\pm 0.04}$ & $0.51_{\pm 0.06}$ & $0.66_{\pm 0.06}$\\
& & Incorrect & $0.46_{\pm 0.11}$ & $0.87_{\pm 0.08}$ & $0.32_{\pm 0.11}$ & $0.67_{\pm 0.09}$\\
\cmidrule{2-7}
& \multirow{2}{*}{high} & Correct & $0.80_{\pm 0.06}$ & $0.87_{\pm 0.05}$ & $0.68_{\pm 0.08}$ & $0.67_{\pm 0.06}$\\
& & Incorrect & $0.68_{\pm 0.14}$ & $0.84_{\pm 0.11}$ & $0.38_{\pm 0.19}$ & $0.67_{\pm 0.11}$\\
        \bottomrule
    \end{tabular}
    }
    \caption{Recreation of \cref{tab:right-wrong-table} using an LLM Judge (GPT-4o with Human references) as a proxy correctness label to ablate the influence of reasoning on LLM Judge performance. Our main result is preserved across reasoning levels.}
    \label{tab:reasoning-ablation-table}
\end{table*}

%% file: judge_prompts.tex
\subsection{Single, Reference, Turn One}
\noindent\textbf{System Prompt:}

\noindent You are a helpful assistant.

\noindent\textbf{Prompt Template:}
\begin{lstlisting}

[Instruction]
Please act as an impartial judge and evaluate the correctness of the response provided by an AI assistant to the user question. You will be given a reference correct answer and the assistant's answer. The reference answer is always correct. Begin you evaluation by comparing the assistant's answer to the correct answer. Identify any mistakes. Be as objective as possible. After providing your explanation, you must rule whether the response is correct or incorrect. Output your final verdict by strictly following this format: "[[Correct]]" if the answer is correct and "[[Incorrect]]" is the answer is incorrect.

[Question]
{question}

[The Start of Reference Answer]
{ref_answer_1}
[The End of Reference Answer]

[The Start of Assistant's Answer]
{answer}
[The End of Assistant's Answer]

\end{lstlisting}

\subsection{Single, Reference, Turn Two}
\noindent\textbf{System Prompt:}

\noindent Please act as an impartial judge and evaluate the correctness of the response provided by an AI assistant to the user question. You will be given a reference correct answer and the assistant's answer. The reference answer is always correct. You should only focus on the assitants answer to the second question. Begin you evaluation by comparing the assistant's answer to the correct answer. Identify any mistakes. Be as objective as possible. After providing your explanation, you must rule whether the response is correct or incorrect. Output your final verdict by strictly following this format: "[[Correct]]" if the answer is correct and "[[Incorrect]]" is the answer is incorrect.

\noindent \textbf{Prompt Template:}
\begin{lstlisting}

<|The Start of Reference Answer|>

### User:
{question_1}

### Reference answer:
{ref_answer_1}

### User:
{question_2}

### Reference answer:
{ref_answer_2}

<|The End of Reference Answer|>

<|The Start of Assistant A's Conversation with User|>

### User:
{question_1}

### Assistant A:
{answer_1}

### User:
{question_2}

### Assistant A:
{answer_2}

<|The End of Assistant A's Conversation with User|>
\end{lstlisting}

\subsection{Single, No Reference, Turn One}
\noindent\textbf{System Prompt:}

\noindent You are a helpful assistant.

\noindent\textbf{Prompt Template:}
\begin{lstlisting}

[Instruction]
Please act as an impartial judge and evaluate the correctness of the response provided by an AI assistant to the user question. You will only be given the assistant's answer. Identify any mistakes. Be as objective as possible. After providing your analysis, you must rule whether the response is correct or incorrect. Output your final verdict by strictly following this format: "[[Correct]]" if the answer is correct and "[[Incorrect]]" is the answer is incorrect.

[Question]
{question}

[The Start of Assistant's Answer]
{answer}
[The End of Assistant's Answer]

\end{lstlisting}

\subsection{Single, No Reference, Turn Two}
\noindent\textbf{System Prompt:}

\noindent Please act as an impartial judge and evaluate the correctness of the response provided by an AI assistant to the user question. You will only be given the assistant's answer. You should only focus on the assitant's answer to the second question. Identify any mistakes. Be as objective as possible. After providing your analysis, you must rule whether the response is correct or incorrect. Output your final verdict by strictly following this format: "[[Correct]]" if the answer is correct and "[[Incorrect]]" is the answer is incorrect.

\noindent\textbf{Prompt Template:}
\begin{lstlisting}

<|The Start of Assistant A's Conversation with User|>

### User:
{question_1}

### Assistant A:
{answer_1}

### User:
{question_2}

### Assistant A:
{answer_2}

<|The End of Assistant A's Conversation with User|>
\end{lstlisting}

\subsection{Pairwise, Reference, Turn One}

\noindent\textbf{System Prompt:}

\noindent Please act as an impartial judge and evaluate the correctness of the responses provided by two AI assistants to the user question displayed below. Your evaluation should consider only correctness. You will be given a reference answer, assistant A's answer, and assistant B's answer. The reference answer is always correct. Your job is to evaluate which assistant's answer is more correct. Begin your evaluation by comparing both assistants' answers with the reference answer. Identify any mistakes. Avoid any position biases and ensure that the order in which the responses were presented does not influence your decision. Do not allow the length of the responses to influence your evaluation. Do not favor certain names of the assistants. Be as objective as possible. After providing your explanation, output your final verdict by strictly following this format: "[[A]]" if assistant A is more correct, and "[[B]]" if assistant B is more correct.

\noindent\textbf{Prompt Template:}
\begin{lstlisting}

[User Question]
{question}

[The Start of Reference Answer]
{ref_answer_1}
[The End of Reference Answer]

[The Start of Assistant A's Answer]
{answer_a}
[The End of Assistant A's Answer]

[The Start of Assistant B's Answer]
{answer_b}
[The End of Assistant B's Answer]
\end{lstlisting}

\subsection{Pairwise, Reference, Turn Two}
\noindent\textbf{System Prompt:}

\noindent Please act as an impartial judge and evaluate the quality of the responses provided by two AI assistants to the user questions. Your evaluation should consider only correctness. You will be given reference answers, the assistant A's answers, the assistant B's answers. The reference answer is always correct. Your job is to determine which assistant provides a more correct answer to the second user question. Begin your evaluation by comparing both assistants' answers with the reference answers. Identify any mistakes. Avoid any position biases and ensure that the order in which the responses were presented does not influence your decision. Do not allow the length of the responses to influence your evaluation. Do not favor certain names of the assistants. Be as objective as possible. After providing your explanation, output your final verdict by strictly following this format: "[[A]]" if assistant A is more correct, and "[[B]]" if assistant B is more correct.

\noindent\textbf{Prompt Template:}
\begin{lstlisting}

<|The Start of Reference Answer|>

### User:
{question_1}

### Reference answer:
{ref_answer_1}

### User:
{question_2}

### Reference answer:
{ref_answer_2}

<|The End of Reference Answer|>

<|The Start of Assistant A's Conversation with User|>

### User:
{question_1}

### Assistant A:
{answer_a_1}

### User:
{question_2}

### Assistant A:
{answer_a_2}

<|The End of Assistant A's Conversation with User|>

<|The Start of Assistant B's Conversation with User|>

### User:
{question_1}

### Assistant B:
{answer_b_1}

### User:
{question_2}

### Assistant B:
{answer_b_2}

<|The End of Assistant B's Conversation with User|>
\end{lstlisting}

\subsection{Pairwise, No Reference, Turn One}
\noindent\textbf{System Prompt:}

\noindent Please act as an impartial judge and evaluate the correctness of the responses provided by two AI assistants to the user question displayed below. You should choose the assistant that gives the more correct response. Your evaluation should only consider correctness. Begin your evaluation by comparing the two responses and provide a short explanation. Avoid any position biases and ensure that the order in which the responses were presented does not influence your decision. Do not allow the length of the responses to influence your evaluation. Do not favor certain names of the assistants. Be as objective as possible. After providing your explanation, output your final verdict by strictly following this format: "[[A]]" if assistant A is more correct, and "[[B]]" if assistant B is more correct.

\noindent\textbf{Prompt Template:}
\noindent\begin{lstlisting}

[User Question]
{question}

[The Start of Assistant A's Answer]
{answer_a}
[The End of Assistant A's Answer]

[The Start of Assistant B's Answer]
{answer_b}
[The End of Assistant B's Answer]
\end{lstlisting}

\subsection{Pairwise, No Reference, Turn Two }
\noindent\textbf{System Prompt:}

\noindent Please act as an impartial judge and evaluate the correctness of the responses provided by two AI assistants to the user questions. You should choose the assistant that gives the more correct response. Your evaluation should only consider correctness. You should focus on who provides a more correct answer to the second user question. Begin your evaluation by comparing the responses of the two assistants and provide a short explanation. Avoid any position biases and ensure that the order in which the responses were presented does not influence your decision. Do not allow the length of the responses to influence your evaluation. Do not favor certain names of the assistants. Be as objective as possible. After providing your explanation, output your final verdict by strictly following this format: "[[A]]" if assistant A is more correct, and "[[B]]" if assistant B is more correct.

\noindent\textbf{Prompt Template:}
\begin{lstlisting}

<|The Start of Assistant A's Conversation with User|>

### User:
{question_1}

### Assistant A:
{answer_a_1}

### User:
{question_2}

### Assistant A:
{answer_a_2}

<|The End of Assistant A's Conversation with User|>

<|The Start of Assistant B's Conversation with User|>

### User:
{question_1}

### Assistant B:
{answer_b_1}

### User:
{question_2}

### Assistant B:
{answer_b_2}

<|The End of Assistant B's Conversation with User|>
\end{lstlisting}